\documentclass{article}

\usepackage[preprint]{icml2026}

\usepackage{amsmath,amssymb,amsthm,mathtools,bm}
\usepackage{booktabs}
\usepackage{graphicx}
\usepackage{algorithm}
\usepackage{hyperref}
\usepackage{microtype}
\usepackage{xcolor}
\hypersetup{hidelinks}

\providecommand{\State}{\STATE}
\providecommand{\For}[1]{\FOR{#1}}
\providecommand{\EndFor}{\ENDFOR}
\providecommand{\Require}{\REQUIRE}

\newtheorem{definition}{Definition}

\newcommand{\SCAR}{\textsc{Scar}}
\newcommand{\LP}{\mathrm{LP}}
\newcommand{\Conn}{\mathrm{Conn}}
\newcommand{\Red}{\mathrm{Red}}
\newcommand{\Protect}{\mathrm{Protect}}

\providecommand{\MainSparsityPct}{50}
\providecommand{\PPLUnpruned}{--}
\providecommand{\PPLWandaChan}{--}
\providecommand{\PPLSparseGPTChan}{--}

\providecommand{\PPLSCARLP}{--}
\providecommand{\PPLSCARProt}{--}
\providecommand{\SupernodeRhoPct}{1}
\providecommand{\TopRhoMassMedianPct}{--}
\providecommand{\TopRhoMassMinPct}{--}
\providecommand{\TopRhoMassMaxPct}{--}
\providecommand{\OLMoTopRhoMassFinalPct}{--}
\providecommand{\OLMoTopRhoMassInitPct}{--}
\providecommand{\OLMoJaccardInitFinalPct}{--}
\providecommand{\OLMoJaccardMidFinalPct}{--}
\providecommand{\OLMoMaxMeanInit}{--}
\providecommand{\OLMoMaxMeanFinal}{--}
\providecommand{\OLMoPPLUnpruned}{--}
\providecommand{\OLMoPPLSCARLP}{--}
\providecommand{\OLMoPPLWandaChan}{--}
\providecommand{\OLMoPPLMagnitude}{--}

\providecommand{\MainSparsityPct}{}
\renewcommand{\MainSparsityPct}{50}

\providecommand{\PPLUnpruned}{}
\renewcommand{\PPLUnpruned}{6.2}
\providecommand{\PPLWandaChan}{}
\renewcommand{\PPLWandaChan}{989.2}
\providecommand{\PPLSparseGPTChan}{}
\renewcommand{\PPLSparseGPTChan}{5256.3}

\providecommand{\PPLSCARLP}{}
\renewcommand{\PPLSCARLP}{55.8}
\providecommand{\PPLSCARProt}{}
\renewcommand{\PPLSCARProt}{54.8}
\providecommand{\PPLSCARHaloProtect}{}
\renewcommand{\PPLSCARHaloProtect}{42.6}
\providecommand{\AccSCARHaloProtect}{}
\renewcommand{\AccSCARHaloProtect}{45.6}

\providecommand{\SupernodeRhoPct}{}
\renewcommand{\SupernodeRhoPct}{1}
\providecommand{\TopRhoMassMedianPct}{}
\renewcommand{\TopRhoMassMedianPct}{58.7}
\providecommand{\TopRhoMassMinPct}{}
\renewcommand{\TopRhoMassMinPct}{33.0}
\providecommand{\TopRhoMassMaxPct}{}
\renewcommand{\TopRhoMassMaxPct}{86.1}

\providecommand{\OLMoTopRhoMassFinalPct}{}
\renewcommand{\OLMoTopRhoMassFinalPct}{71.7}
\providecommand{\OLMoTopRhoMassInitPct}{}
\renewcommand{\OLMoTopRhoMassInitPct}{18.6}
\providecommand{\OLMoJaccardInitFinalPct}{}
\renewcommand{\OLMoJaccardInitFinalPct}{0.5}
\providecommand{\OLMoJaccardMidFinalPct}{}
\renewcommand{\OLMoJaccardMidFinalPct}{46.7}
\providecommand{\OLMoMaxMeanInit}{}
\renewcommand{\OLMoMaxMeanInit}{77}
\providecommand{\OLMoMaxMeanFinal}{}
\renewcommand{\OLMoMaxMeanFinal}{3690}
\providecommand{\OLMoPPLUnpruned}{}
\renewcommand{\OLMoPPLUnpruned}{6.03}
\providecommand{\OLMoPPLSCARLP}{}
\renewcommand{\OLMoPPLSCARLP}{15.43}
\providecommand{\OLMoPPLWandaChan}{}
\renewcommand{\OLMoPPLWandaChan}{15.20}
\providecommand{\OLMoPPLMagnitude}{}
\renewcommand{\OLMoPPLMagnitude}{134.08}

\newcommand{\TODO}[1]{}
\newcommand{\NEWFIG}[1]{}
\newcommand{\NEWEXP}[1]{}

\newcommand{\ReleaseCodeURL}{\url{https://github.com/KempnerInstitute/nodelens}}
\newcommand{\ReleaseArtifactsURL}{\url{https://huggingface.co/datasets/hsafaai/supernodes-scar-artifacts}}

\icmltitlerunning{Loss-Critical Hubs in LLM FFNs}

\begin{document}

\twocolumn[
\icmltitle{Supernodes and Halos: Loss-Critical Hubs in LLM Feed-Forward Layers}

\begin{center}
{\large\bfseries Audrey Cherilyn, Houman Safaai\textsuperscript{1}}\par\vspace{0.08in}
{\normalsize Kempner Institute at Harvard University}
\end{center}

\icmlkeywords{large language models, pruning, loss sensitivity, Fisher information, structured sparsity}

\vskip 0.3in
]

\footnotetext[1]{Correspondence: \texttt{houman\_safaai@harvard.edu}.}

\makeatletter
\global\icml@noticeprintedtrue
\makeatother

\begin{abstract}
We study the organization of channel-level importance in transformer feed-forward networks (FFNs). Using a Fisher-style loss proxy (LP) based on activation--gradient second moments, we show that loss sensitivity is concentrated in a small set of channels within each layer. In Llama-3.1-8B, the top \SupernodeRhoPct\% of channels per layer accounts for a median of \TopRhoMassMedianPct\% of LP mass (range: \TopRhoMassMinPct--\TopRhoMassMaxPct\%). We call these loss-critical channels \textbf{supernodes}. Although FFN layers also contain strong activation outliers, LP-defined supernodes overlap only weakly with activation-defined outliers and are not explained by activation power or weight norms alone. Around this core, we find a weaker but consistent halo structure: same-layer channels with overlapping write support show stronger redundancy with the protected core, and next-layer channels that read from supernode-supported coordinates respond more strongly to support ablation. We use one-shot structured FFN pruning as a diagnostic test of this organization. At 50\% FFN sparsity, baselines that prune many supernodes degrade sharply, whereas \SCAR{} variants explicitly protect the supernode core; \SCAR{}-Prot reaches perplexity \PPLSCARProt{}, and a separate \SCAR{}-HaloProtect run reaches \PPLSCARHaloProtect{}, compared with \PPLWandaChan{} for Wanda-channel. The LP-concentration pattern appears across Mistral-7B, Llama-2-7B, and Qwen2-7B, remains visible in targeted Llama-3.1-70B experiments, and increases during OLMo-2-7B pretraining. At higher sparsity, hard read-halo protection further reduces collapse when \SCAR{}-LP begins to fail. These results suggest that LLM FFNs develop a small learned core of loss-critical channels, and that preserving this core and its supporting halo is important for reliable structured pruning.
\end{abstract}

\section{Introduction}
\label{sec:intro}

Large language models are not homogeneous computations. A growing body of evidence shows that a small number of internal components can carry disproportionate functional importance: massive activation coordinates can shape attention behavior \citep{sun2024massive}, and individual ``super weights'' can strongly affect model outputs \citep{yu2024super}. These results raise a natural question at a more structured level: do transformer FFNs contain a small set of whole channels that are unusually important for loss?

This question matters both scientifically and practically. FFN channels are coherent computational units \citep{geva2021keyvalue,geva2022promoting,dai2022knowledge}: each channel reads from the residual stream through its up and gate projections, produces one intermediate activation, and writes back through one column of the down projection. Unlike individual weights, channels can be ablated and pruned as deployable structures. If their importance is highly concentrated, then structured pruning may fail not because most channels are indispensable, but because pruning methods eventually remove a small protected core.

\begin{figure*}[!t]
    \centering
    \includegraphics[width=\textwidth]{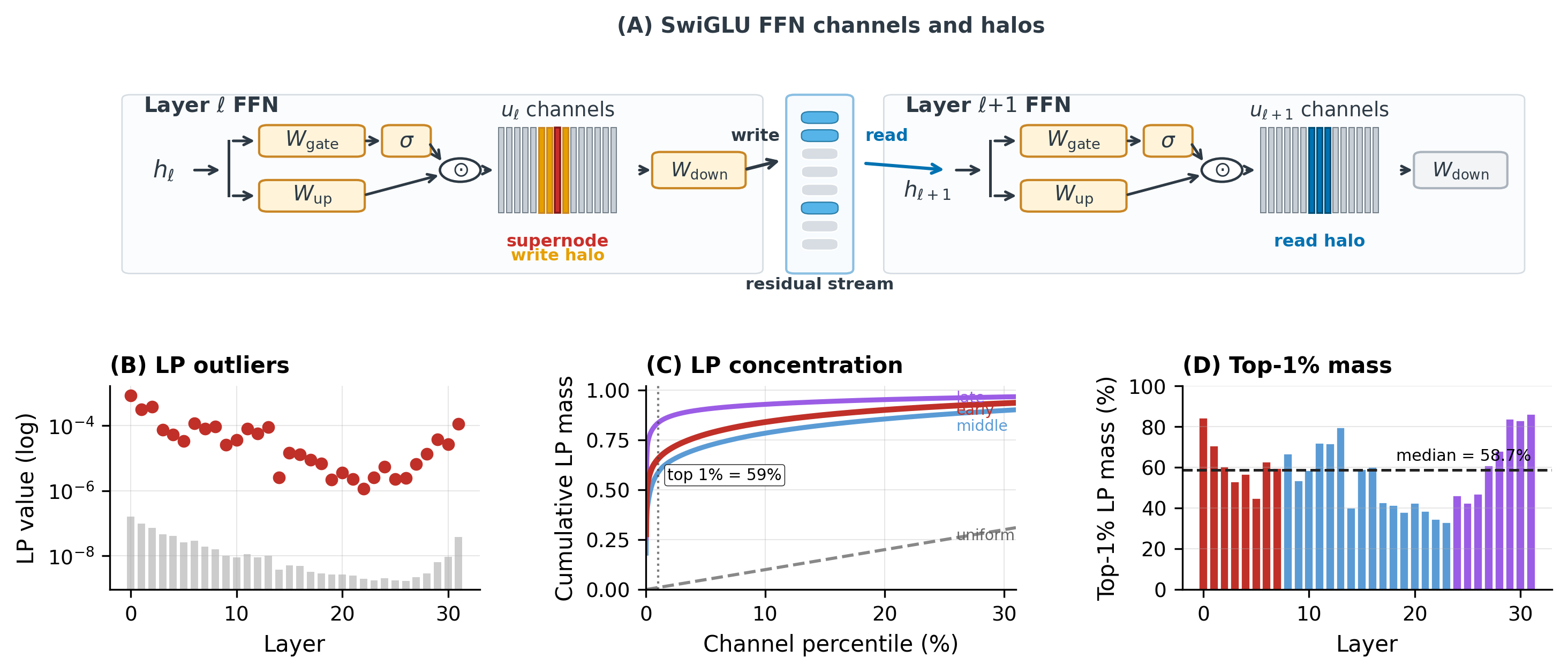}
    \caption{\textbf{Supernodes, halos, and LP concentration.}
\textbf{(A)} SwiGLU FFN channels are intermediate coordinates that read through \(W_{\rm gate}\) and \(W_{\rm up}\), write through \(W_{\rm down}\), and couple adjacent layers through residual-stream coordinates; red bars are supernodes, orange bars are write halos, blue residual coordinates form the support \(\mathcal{S}_\ell\), and blue layer-\((\ell{+}1)\) bars are read halos.
\textbf{(B--D)} LP is highly concentrated in a small per-layer core, with the strongest concentration in early and late layers.}
    \label{fig:conceptual}
    \vspace{-0.7em}
\end{figure*}

Figure~\ref{fig:conceptual} gives the high-level picture. A small set of FFN channels, shown in red, writes into a sparse residual support $\mathcal{S}_\ell$. Channels in the next layer that read strongly from this support form a read halo, while same-layer channels that write to the same support form a write halo. The remaining panels show why the red core deserves special treatment: LP mass is concentrated sharply in every layer, especially in early and late layers.

Structured pruning provides a useful probe of this hypothesis. Removing whole FFN channels is attractive for deployable compression \citep{ma2023llmpruner,an2024flap,yin2024owl,guo2025slimllm}, but it is more brittle than unstructured pruning \citep{frantar2023sparsegpt,sun2024wanda}. The reason is simple: unstructured pruning can remove isolated low-saliency weights, whereas channel pruning removes complete computational units. If a few channels are loss-critical, then an otherwise reasonable structured mask can become highly damaging once it touches this core.

We study the intermediate channels of SwiGLU FFNs, which account for a large fraction of the parameters in modern decoder LLMs \citep{touvron2023llama2,dubey2024llama3,jiang2023mistral,yang2024qwen2}. Our main diagnostic is a Fisher-style loss proxy (LP) that combines each channel's activation with the gradient signal projected onto its write direction. The resulting channel distribution is extremely skewed. In Llama-3.1-8B, the top 1\% of channels per layer captures between \TopRhoMassMinPct\% and \TopRhoMassMaxPct\% of the total LP mass, with median \TopRhoMassMedianPct\%. We call these top-LP channels \textbf{supernodes}.

Supernodes are not merely large activations or large-norm weights. Their LP values can exceed the layer mean by roughly three orders of magnitude, and targeted ablations show that they are functionally critical. Forcing a pruning mask to remove supernodes drives perplexity from \PPLUnpruned{} to more than $8\times 10^5$, even at only \MainSparsityPct\% sparsity, whereas removing the same number of non-supernode channels is far less damaging (Appendix Table~\ref{tab:supernode-control}). This identifies a concrete failure mode for aggressive structured pruning: pruning methods can tolerate many non-core removals, but fail sharply when they hit the small loss-critical core.

Because LP is estimated from calibration data, we distinguish two types of stability. The phenomenon itself is stable: extreme LP concentration appears across calibration domains. The specific channel identities are less stable and depend on the calibration distribution. This implies a practical rule: \SCAR{} should be calibrated on data representative of the intended deployment domain.

After identifying the supernode core, we ask whether the surrounding non-core channels contain useful structure. We find a cross-layer \textbf{read halo}: next-layer FFN channels with disproportionate input-weight mass on supernode-emphasized residual coordinates show elevated within-set activation redundancy and respond preferentially to causal perturbations of the supernode support, consistent with the read halo \emph{echoing} the supernode signal into the next layer. A same-layer write halo recapitulates this organization in weaker form. The halo structure is secondary to the supernode core for pruning quality, but it is the natural mechanistic completion of the picture: supernodes do not act in isolation but generate downstream redundancy.

We validate this structure through one-shot structured FFN pruning. At 50\% FFN channel sparsity on Llama-3.1-8B, the baselines in Table~\ref{tab:main} remove roughly 20--42\% of supernodes and degrade to perplexities of several hundred to several thousand. In contrast, \SCAR{} variants protect all supernodes and remain substantially more stable: \SCAR{}-Prot reaches perplexity \PPLSCARProt{}, and the separate main-pipeline \SCAR{}-HaloProtect run reaches \PPLSCARHaloProtect{}, compared with \PPLWandaChan{} for channel-adapted Wanda. Across methods, supernode hit-rate is strongly associated with perplexity, and controlled random-mask sweeps show a clear dose--response relationship.

\paragraph{Contributions.}
The paper makes four contributions. First, we define a Fisher-style channel loss proxy whose mass concentrates in per-layer supernodes and is not reducible to activation, curvature, or their factorization. Second, we identify direct supernode removal as a major failure mode of one-shot structured FFN pruning. Third, we characterize a cross-layer read-halo structure: channels in layer $\ell{+}1$ that read from the supernode-supported residual coordinates of layer $\ell$ behave partly redundantly at moderate sparsity, and become load-bearing at high sparsity. Fourth, we validate the pattern across Llama-3.1-8B, Llama-2-7B, Mistral-7B, Qwen2-7B, targeted Llama-3.1-70B experiments, and ten OLMo-2-7B pretraining checkpoints.

\begin{figure*}[!t]
    \centering
    \includegraphics[width=\textwidth]{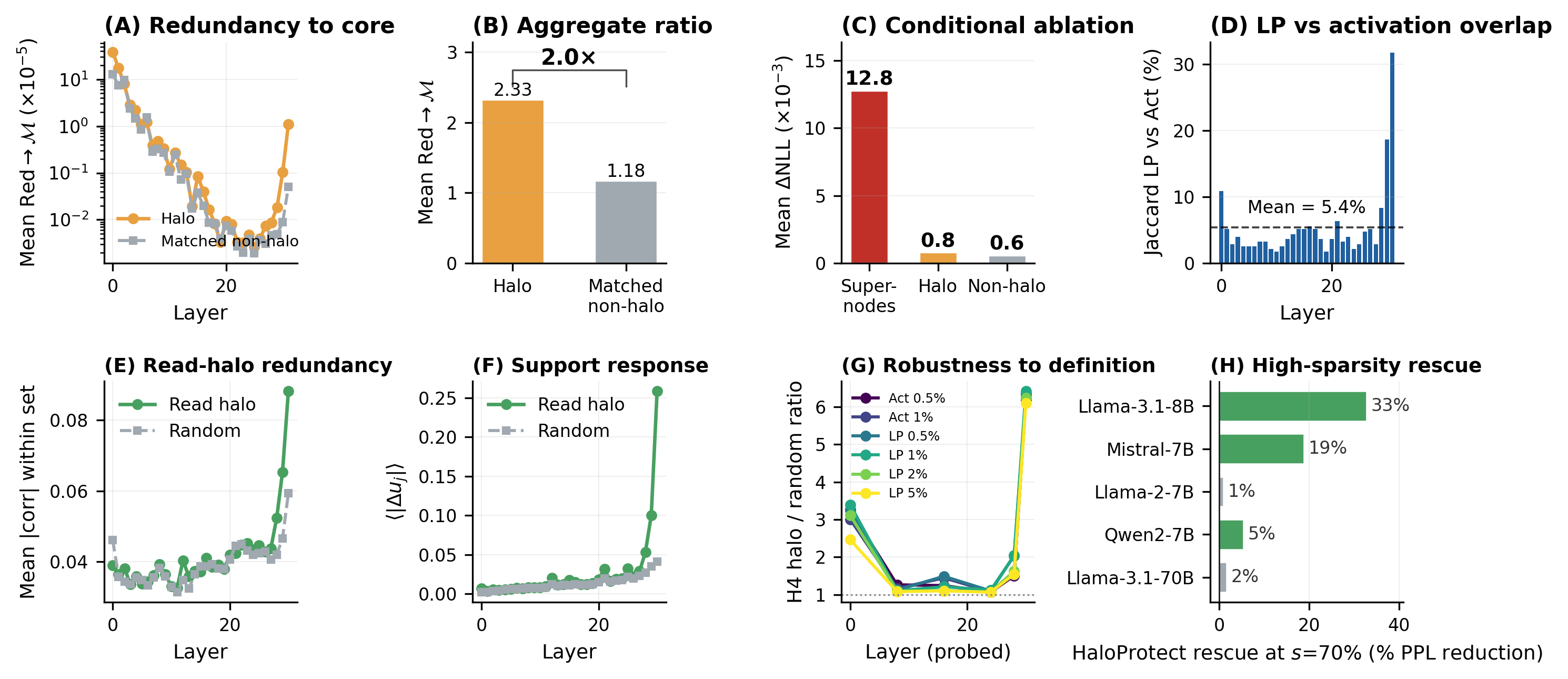}
    \caption{\textbf{Halo diagnostics and LP-vs-activation controls.}
    \textbf{(A--C)} Same-layer write halos show modest redundancy, while direct supernode removal dominates ablation cost.
    \textbf{(D)} LP supernodes overlap weakly with activation top channels.
    \textbf{(E,F)} Cross-layer read halos show stronger within-set redundancy and larger response to support ablation.
    \textbf{(G,H)} The read-halo signal is robust to the supernode definition and helps most under aggressive sparsity.}
    \label{fig:halo-structure}
\end{figure*}

\section{Background and Related Work}
\label{sec:related}

Supernodes sit between two previously studied forms of rare but important structure in LLMs. They are more structured than individual critical weights, but they are not simply activation outliers. They are whole FFN channels whose importance is defined through loss sensitivity.

\paragraph{FFN channels.}
\label{sec:prelim}

Modern decoder LLMs such as Llama, Mistral, and Qwen commonly use SwiGLU-style feed-forward layers \citep{touvron2023llama2,dubey2024llama3,jiang2023mistral,yang2024qwen2}.
For layer $\ell$,
\begin{align}
    z &= W_{\text{up}} h, \quad g = W_{\text{gate}} h, \\
    u &= \sigma(g) \odot z \in \mathbb{R}^m, \quad y = W_{\text{down}} u \in \mathbb{R}^d,
\end{align}
where $h \in \mathbb{R}^d$ is the input hidden state, $\sigma$ is SiLU, and \(m\) is typically several times larger than \(d\). For example, Llama-3.1-8B has \(d=4096\) and \(m=14336 \approx 3.5d\).

A channel \(i \in \{1,\ldots,m\}\) is the \(i\)-th coordinate of the intermediate representation \(u\), together with row \(i\) of \(W_{\text{up}}\), row \(i\) of \(W_{\text{gate}}\), and column \(i\) of \(W_{\text{down}}\). We write
\[
    v_i = W_{\text{down}}[:, i] \in \mathbb{R}^d
\]
for the channel's write direction. The rows \(W_{\text{up}}[i,:]\) and \(W_{\text{gate}}[i,:]\) describe what the channel reads from the residual stream, while \(v_i\) describes where it writes. Our halo definitions use both sides: write halos are defined within a layer by overlap with the supernode write support, while read coupling is defined across layers by overlap between next-layer read weights and that same support.

\paragraph{Outlier phenomena in LLMs.}
Rare internal components can have large effects on LLM behavior. At the activation level, \citet{sun2024massive} show that a small number of activation coordinates can behave like implicit bias terms and create attention sinks. Activation outliers also affect compression: LLM.int8() isolates outlier dimensions in higher precision \citep{dettmers2022llmint8}, SmoothQuant shifts scale from activations to weights to reduce activation-outlier difficulty \citep{xiao2022smoothquant}, and \citet{bondarenko2024quant} characterize how attention-head no-op states create the outlier structure that compression must accommodate. At the parameter level, \citet{yu2024super} identify individual ``super weights'' whose removal can destroy model output and induce corresponding super activations. Mechanistic-interpretability work at scale \citep{templeton2024scaling} also reports that small numbers of features carry disproportionate functional importance.

Our work asks whether an analogous phenomenon exists at the level of whole FFN channels. This channel-level object differs from prior outlier units in two ways. First, a channel is a structured component spanning rows and columns across the FFN projections. Second, we identify channels by loss sensitivity rather than activation magnitude alone. We then test their role directly through ablation and structured pruning.

\paragraph{Structured pruning of LLMs.}
Pruning is both a compression method and a diagnostic for functional importance. Wanda \citep{sun2024wanda} and SparseGPT \citep{frantar2023sparsegpt} are strong one-shot unstructured methods, but unstructured sparsity usually requires specialized kernels to yield hardware gains. Structured methods remove larger units: Group Fisher Pruning scores parameter groups with Fisher information \citep{liu2021group}, the Optimal BERT Surgeon \citep{kurtic2022obert} extends second-order pruning to large transformers, LLM-Pruner performs structural pruning with LoRA recovery \citep{ma2023llmpruner}, OWL adds outlier-aware non-uniform sparsity allocation \citep{yin2024owl}, and OATS decomposes weights into sparse and low-rank terms \citep{oats}. \citet{frantar2024scaling} study scaling laws for sparsely-connected foundation models, and \citet{voita2019heads} earlier showed an analogous ``few units carry most of the work'' phenomenon for attention heads.

We focus on the harder one-shot setting in which whole FFN channels are removed without recovery fine-tuning. Our contribution is not Fisher scoring by itself. Rather, we show that a Fisher-style channel score reveals an extremely concentrated supernode core, and that direct removal of this core explains a major failure mode of aggressive structured pruning.

\section{Supernode Characterization}
\label{sec:supernodes}

We now define channel-level loss sensitivity and characterize the structure it reveals.

\paragraph{Loss proxy.}
Masking channel \(i\) sets \(u_i=0\), which induces the FFN output perturbation \(\Delta y=-u_i v_i\), where \(v_i=W_{\text{down}}[:,i]\). Under a Fisher/Gauss--Newton approximation, the expected loss increase from this perturbation is
\begin{equation}
    \mathbb{E}[\Delta \mathcal{L}] \approx \frac{1}{2} \mathbb{E}\left[ u_i^2 \cdot (v_i^\top g_y)^2 \right],
\end{equation}
where \(g_y=\nabla_y \mathcal{L}\) is the loss gradient at the FFN output. Let
\[
    s_i = v_i^\top g_y = (\nabla_u \mathcal{L})_i
\]
be the gradient signal projected onto channel \(i\)'s write direction. The resulting channel loss proxy is

\begin{equation}
    \LP_i = \frac{1}{2} \mathbb{E}_{x \sim \mathcal{D}} \left[ (u_i \cdot s_i)^2 \right].
    \label{eq:loss-proxy}
\end{equation}

This proxy captures the joint second moment of activation and projected gradient. A large \(\LP_i\) can come from large activations, large projected gradients, or strong sample-level alignment between the two. It therefore measures loss-sensitive channel importance rather than magnitude alone. In practice, it is efficient to compute: the forward pass records \(u_i\), the backward pass records \(s_i=(W_{\text{down}}^\top g_y)_i\), and the estimate accumulates \(\frac{1}{2}(u_i s_i)^2\) over calibration tokens. No per-channel backward passes are required.

Appendix Fig.~\ref{fig:lp-proxy-diagnostics} validates \(\LP\) through top-tail ablation, and Appendix Table~\ref{tab:mechanism-controls} shows that the score is not explained away by activation power or weight norms.

\paragraph{Supernode definition.}
We define supernodes within each layer rather than globally across the model. This makes scores comparable across layers with different scales and gives every layer a small candidate core.

\begin{definition}[Supernodes]
\label{def:supernodes}
For layer \(\ell\) with \(m\) intermediate channels and loss proxy scores \(\{\LP_i\}_{i=1}^m\), let \(\operatorname{Top}_{\rho}(\{\LP_i\})\) denote the \(\lceil \rho m\rceil\) channels with the largest LP values. The \emph{supernode set} is
\[
    \mathcal{M}_\ell = \operatorname{Top}_{\rho}(\{\LP_i\}_{i=1}^m).
\]
Unless stated otherwise, we use \(\rho=0.01\), so supernodes are the top 1\% of FFN channels by LP within each layer.
\end{definition}

\paragraph{Empirical characterization.}
The LP distribution is highly skewed. Across all layers of Llama-3.1-8B, the top \(\rho=1\%\) of channels by \(\LP_i\) captures between \TopRhoMassMinPct\% and \TopRhoMassMaxPct\% of the total proxy mass \(\sum_i \LP_i\), with median \TopRhoMassMedianPct\%. The same concentration pattern appears in Mistral-7B, Llama-2-7B, and Qwen2-7B.

Supernodes are also extreme outliers within their layers. Figure~\ref{fig:conceptual}\textbf{C} compares the maximum LP value in each layer to the layer mean and shows roughly 2.5--4 orders of magnitude of separation, with a median max/mean ratio of about \(2.6\times10^3\). The effect varies with depth and is strongest in early and late layers.

Figure~\ref{fig:conceptual}\textbf{D} shows the same skew through cumulative LP mass: the top few percent of channels accumulates sensitivity far faster than a uniform reference, with the tightest concentration in layers L0--L3 and L28--L31 and a more gradual rise in middle layers.

\paragraph{Joint activation--gradient structure.}
\label{sec:factor-decomposition}

\ifdefined\NeurIPSSubmission
\begin{figure}[!ht]
    \vspace{-0.5em}
    \centering
    \includegraphics[width=0.86\textwidth]{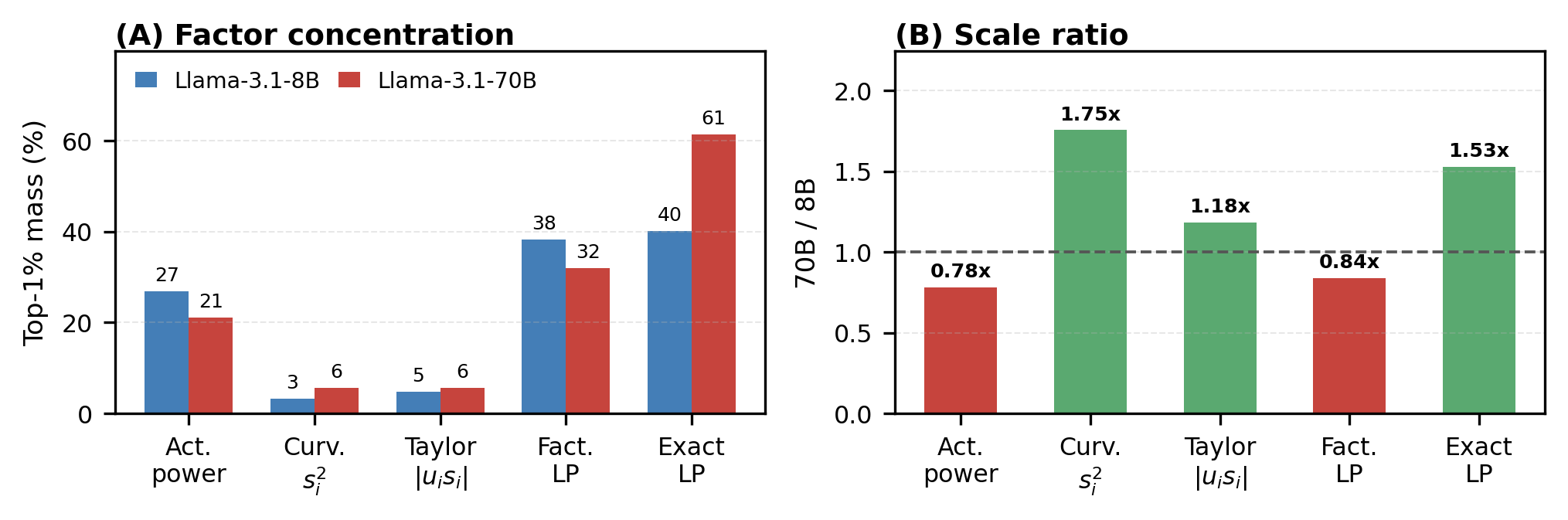}
    \vspace{-0.9em}
    \caption{\textbf{LP factor decomposition: 8B vs 70B.} Activation alone $\mathbb{E}[u_i^2]$ \emph{decreases} with scale; exact LP grows. The gap between exact and factored LP reflects dependence between activation and projected-gradient magnitudes.}
    \label{fig:factor-concentration}
    \vspace{-0.8em}
\end{figure}
\else
\begin{figure*}[!t]
    \centering
    \includegraphics[width=0.92\textwidth]{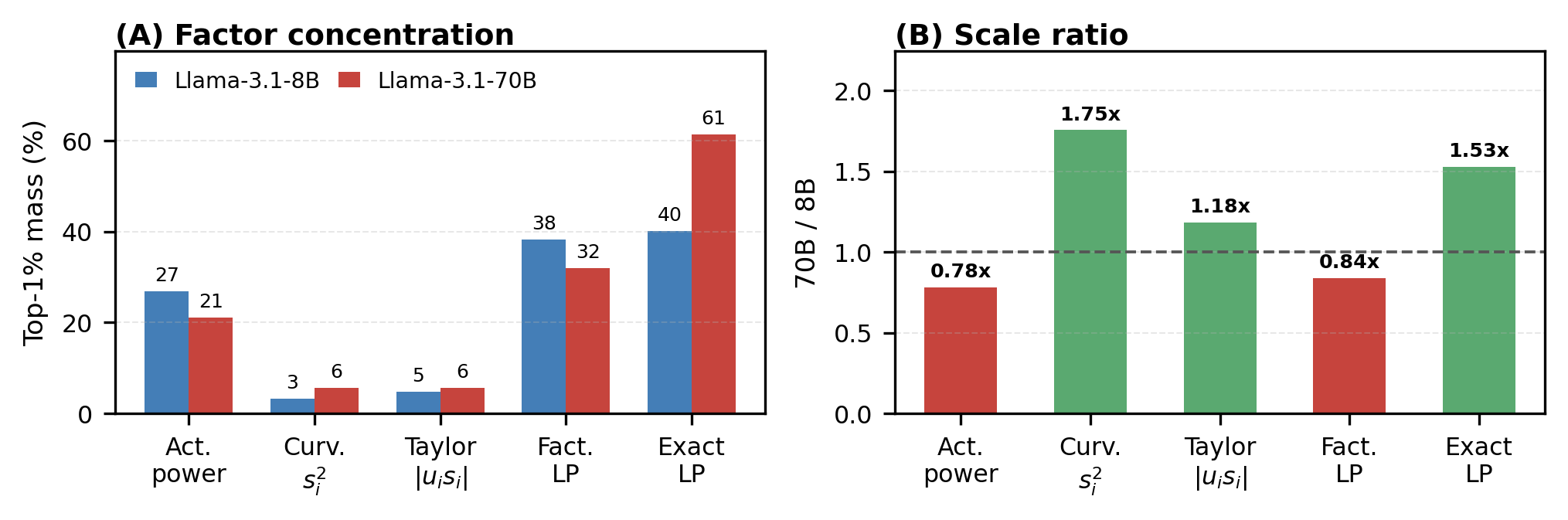}
    \caption{\textbf{LP factor decomposition: 8B vs 70B.} Activation alone $\mathbb{E}[u_i^2]$ \emph{decreases} with scale; exact LP grows. The gap between exact and factored LP reflects dependence between activation and projected-gradient magnitudes.}
    \label{fig:factor-concentration}
\end{figure*}
\fi

LP concentration is not explained by activation magnitude or curvature alone; the gap between LP and its separable approximations \emph{grows} with scale. Under a matched protocol, the top-1\% mass shares are: activation power $\mathbb{E}[u_i^2]$ 27.0\%/21.1\% (8B/70B), curvature $\mathbb{E}[s_i^2]$ 3.2\%/5.6\%, Taylor magnitude $\mathbb{E}[|u_i s_i|]$ 4.9\%/5.8\%, factored $\mathbb{E}[u_i^2]\cdot\mathbb{E}[s_i^2]$ 38.3\%/32.1\%, and exact LP $\mathbb{E}[(u_i s_i)^2]$ 40.2\%/61.4\% (Fig.~\ref{fig:factor-concentration}). The gap between exact and factored LP reflects within-token dependence between activation magnitude and projected-gradient magnitude; it widens from $1.8\,$pp to $29.3\,$pp, suggesting that joint activation--gradient magnitude dependence is a major driver at scale (App.~Table~\ref{tab:mechanism-controls}).

\paragraph{Calibration-domain stability.}
Extreme LP concentration appears consistently across WikiText-2, C4, code, and arXiv calibration data, but the identities of the top channels are domain-dependent. At the 1\% supernode fraction, the mean pairwise Jaccard overlap of LP-defined sets is only $\approx$6\%. This instability is distinct from activation-defined outliers: even within a single domain, the top-1\% LP and activation-defined channel sets overlap only weakly (Fig.~\ref{fig:halo-structure}D), although activation-defined supernodes can still serve as an effective protected-set surrogate when SCAR's pruning score is held fixed (Appendix \S\ref{app:lp-vs-act-supernodes}). Activation-defined supernodes are more stable across domains, while LP-defined supernodes better reflect domain-specific loss sensitivity. This supports calibrating \SCAR{} on data representative of the deployment setting.

\section{Halo Structure Around the Supernode Core}
\label{sec:halos}

\ifdefined\NeurIPSSubmission
\else
\begin{figure*}[!t]
    \centering
    \includegraphics[width=0.92\textwidth]{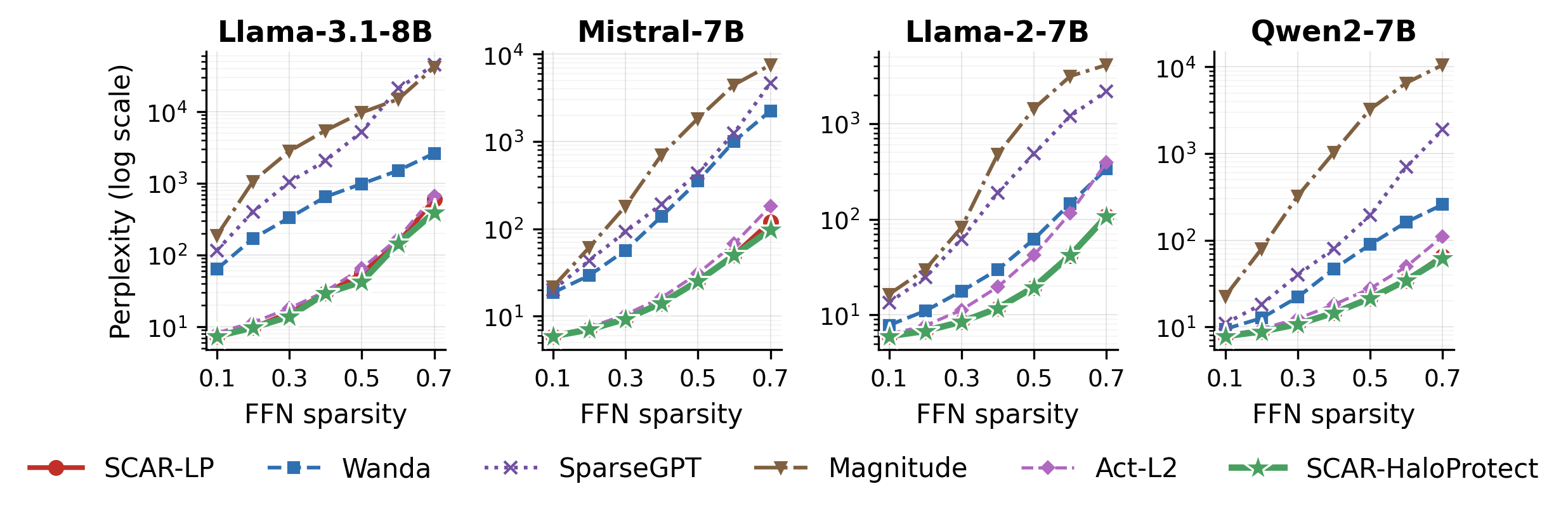}
    \caption{\textbf{Cross-model sparsity--perplexity tradeoff.} SCAR-LP protects the supernode core; \SCAR{}-HaloProtect additionally protects the top 10\% of the read-halo set. HaloProtect tracks SCAR-LP at moderate sparsity and helps most at high sparsity when SCAR-LP begins to collapse.}
    \label{fig:cross-model-curves}
\end{figure*}
\fi

The supernode core explains the strongest pruning failures, but the remaining channels are not structureless. A supernode in layer $\ell$ writes into a small set of residual-stream coordinates; channels that share or read from those coordinates may carry related information. We call these neighborhoods \emph{halos}. The same-layer write halo contains channels in layer $\ell$ that write to the supernode support, while the cross-layer read halo contains channels in layer $\ell{+}1$ that read from that support. The read halo is the more direct test of downstream propagation, so we treat it as the primary halo object.

\paragraph{Shared residual support.}
Both halo definitions are anchored on the residual coordinates emphasized by the supernode core. Let $v_i^{(\ell)} = W^{(\ell)}_{\text{down}}[:, i] \in \mathbb{R}^d$ be the write direction for FFN channel $i$ in layer $\ell$. For supernode set $\mathcal{M}_\ell$, define the aggregated write pattern
\[
    a_\ell = \sum_{s \in \mathcal{M}_\ell} |v_s^{(\ell)}|
\]
and let $\mathcal{S}_\ell=\mathrm{TopK}(a_\ell)$ be the set of its largest coordinates. We use $K=256$, about 6\% of $d=4096$. This support is the only place where supernodes enter the halo construction; the halo scores themselves use static weights and $\mathcal{S}_\ell$.

\paragraph{Read halo.}
The read halo asks which channels in the next layer have gate/up rows concentrated on the coordinates written by the layer-$\ell$ core. For channel $j$ in layer $\ell{+}1$,
\[
    \mathrm{ReadConn}_j =
    \frac{\|W^{(\ell+1)}_{\mathrm{gate}}[j,\mathcal{S}_\ell]\|_1 + \|W^{(\ell+1)}_{\mathrm{up}}[j,\mathcal{S}_\ell]\|_1}{\|W^{(\ell+1)}_{\mathrm{gate}}[j,:]\|_1 + \|W^{(\ell+1)}_{\mathrm{up}}[j,:]\|_1}\,,
\]
the fraction of $j$'s input-weight mass concentrated on $\mathcal{S}_\ell$. The read halo $\mathcal{R}_{\ell+1}$ is the top $\eta_r$ fraction of layer-$(\ell{+}1)$ channels by $\mathrm{ReadConn}$, with default $\eta_r=10\%$. It uses no LP scores from layer $\ell{+}1$; it is defined only by the previous layer's supernode support and the next layer's input-side weights.

Two diagnostics support the read-halo interpretation (Fig.~\ref{fig:halo-structure}E--F; Appendix \S\ref{app:halo-rescue}). First, within-set activation correlation among read-halo channels exceeds a matched random non-halo control by $1.06\times$ on average over valid adjacent layer transitions and by $1.21\times$ in late layers, with the largest single-layer ratio near $2.3\times$. Second, ablating $\mathcal{S}_\ell$ at the input to layer $\ell{+}1$ perturbs read-halo channels more than random channels. The mean $\langle |\Delta u_j|\rangle$ is $1.94\times$ larger overall and $2.80\times$ larger in late layers, with late-layer maxima around $6\times$. This intervention supports a direct dependence of read-halo channels on the supernode support.

The pruning role of read halos depends on the sparsity regime. At moderate sparsity, preferentially pruning read-halo channels is no worse than LP-only pruning and gives a small improvement at 50\%, consistent with partial redundancy while the supernode core and other capacity remain intact. At 70\% sparsity, the ranking flips: read-halo-aware pruning becomes much worse (PPL 1221 vs.\ 629), suggesting that these channels become load-bearing once the remaining network has too little spare capacity.

This motivates \SCAR{}-HaloProtect, which protects supernodes and the top $\tau$ fraction of the read-halo set in each layer while pruning the rest by lowest LP. With $\eta_r=10\%$ and $\tau=10\%$, the extra protected set is about 1\% of channels per valid transition. The main-pipeline 50\% HaloProtect run improves Llama-3.1-8B perplexity (\PPLSCARHaloProtect{} vs.\ \PPLSCARLP{} for \SCAR{}-LP), while the controlled side-pipeline $\tau$ sweep shows that halo protection is nearly neutral at 50\% under that separate loop and helpful at high sparsity. In the side-pipeline sweep on Llama-3.1-8B at $s=70\%$, it reduces perplexity from $628.75$ to $422.73$ (33\%). Across Mistral-7B, Llama-2-7B, Qwen2-7B, and Llama-3.1-70B, the perplexity reduction is largest where SCAR-LP degrades most: 19\% on Mistral, 5\% on Qwen2, and at most 2\% where SCAR-LP already remains stable. The same read-halo pattern is robust to supernode definitions with $\rho \in \{0.5\%,1\%,2\%,5\%\}$ for LP and $\rho\in\{0.5\%,1\%\}$ for activation magnitude (Appendix Fig.~\ref{fig:halo-extended}A).

\paragraph{Same-layer write halo.}
The write halo is the same-layer counterpart to the read halo: it asks which non-supernode channels in layer $\ell$ write into the same residual coordinates as the supernodes. For non-supernode channel $j \in \{1,\ldots,m\} \setminus \mathcal{M}_\ell$,
\[
    \Conn_j = \frac{\sum_{h \in \mathcal{S}_\ell} |v_j^{(\ell)}[h]|}{\|v_j^{(\ell)}\|_1}\,,
\]
the fraction of $j$'s output-weight mass landing on $\mathcal{S}_\ell$. The write halo $\mathcal{H}_\ell$ is the top $\eta=10\%$ of non-supernodes by $\Conn$. Write-halo channels have about $2\times$ higher directed redundancy to the supernode core than matched non-halo channels in loss-relevant contribution space (Fig.~\ref{fig:halo-structure}A--B; derivation in Appendix \S\ref{app:derivations}). The effect is consistent but smaller than the cross-layer read-halo effect, so we use it mainly as a ranking signal among non-core channels.

\section{Validation: Structured Pruning as a Probe}
\label{sec:pruning}

\ifdefined\NeurIPSSubmission
\else
\begin{figure*}[!t]
    \centering
    \includegraphics[width=0.92\textwidth]{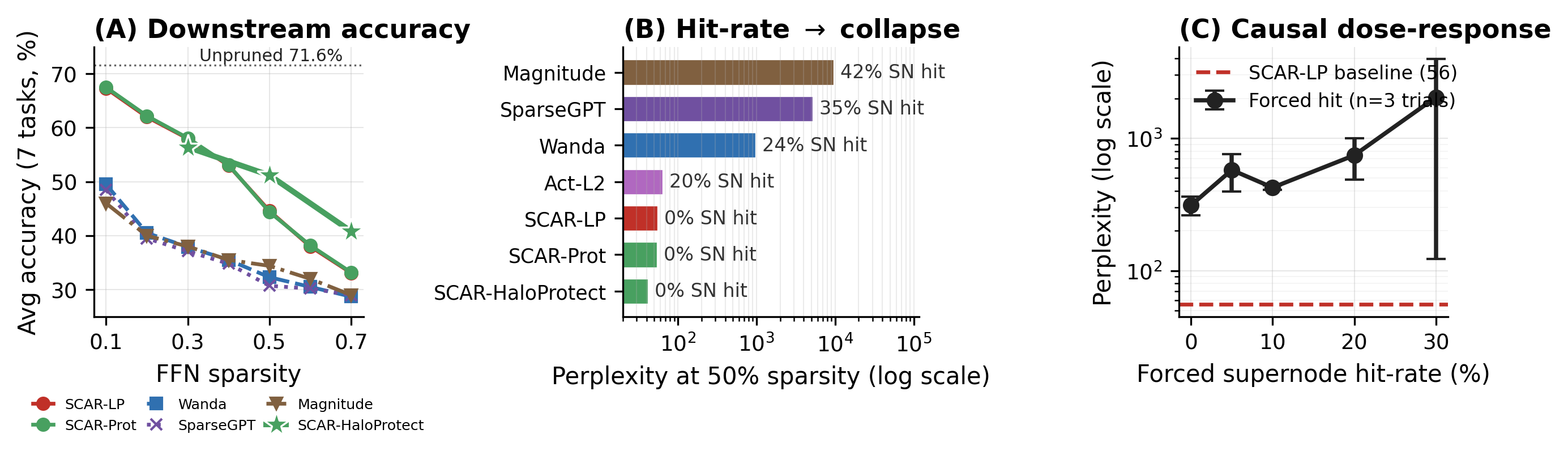}
    \caption{\textbf{Pruning probe and causal evidence on Llama-3.1-8B.}
    \textbf{(A)} SCAR variants retain higher downstream accuracy across sparsities.
    \textbf{(B)} At 50\% sparsity, supernode hit-rate tracks perplexity collapse.
    \textbf{(C)} Forced supernode removal gives a causal dose-response at fixed sparsity.}
    \label{fig:pruning-probe}
\end{figure*}
\fi

If supernodes capture genuine functional organization, pruning strategies that preserve them should be more robust than otherwise similar strategies that remove them.
We use structured FFN channel pruning to test this claim quantitatively, with halo features treated as a secondary refinement after the core is protected.

\ifdefined\NeurIPSSubmission
\begin{figure*}[!t]
    \centering
    \includegraphics[width=\textwidth]{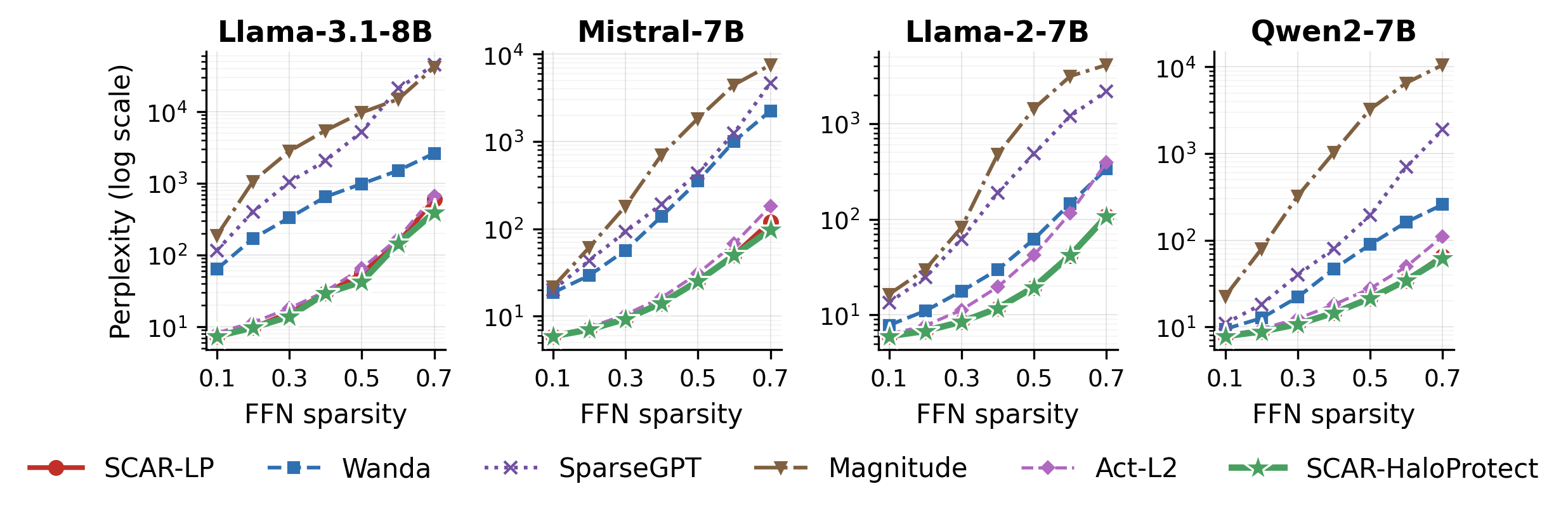}
    \caption{\textbf{Cross-model sparsity--perplexity tradeoff.} SCAR-LP protects the supernode core; \SCAR{}-HaloProtect additionally protects the top 10\% of the read-halo set. HaloProtect tracks SCAR-LP at moderate sparsity and helps most at high sparsity when SCAR-LP begins to collapse.}
    \label{fig:cross-model-curves}
\end{figure*}
\fi

\paragraph{Pruning framework.}
We implement \SCAR{} (\underline{S}upernode-\underline{C}onstrained \underline{A}llocation for \underline{R}esource-aware pruning), a one-shot structured pruning rule that excludes supernodes from the candidate prune set, ranks all remaining FFN channels by a chosen importance score, and selects channels under a global sparsity budget while enforcing per-layer sparsity caps. Halo-aware variants use redundancy and connectivity only after protecting the supernode core.

\paragraph{Scoring variants.}
\SCAR{}-LP prunes the lowest-\(\LP_i\) non-supernode channels and serves as the LP-ranking baseline with explicit core exclusion but no halo refinement.
\SCAR{}-Prot weights \(\LP_i\) by a redundancy-based protection score \(\Protect_i\).
\SCAR{}-Conn further incorporates connectivity, so that redundant halo channels are preferentially removed.
\SCAR{}-HaloProtect is the hard read-halo variant used in the high-sparsity tests: after defining the read halo as the top 10\% of channels by ReadConn, it excludes the top 10\% of that read-halo set from the prune pool, then ranks the remaining channels by LP. The diagnostic variants \SCAR{}-PruneReadHalo, \SCAR{}-ReadHalo, and \SCAR{}-TwoHalo are defined in Appendix~\S\ref{app:halo-rescue}.

The algorithm is detailed in Algorithm~\ref{alg:scar} (Appendix).
Here, the pruning results serve mainly as evidence about the underlying structure.

\paragraph{Experimental setup.}
We use the same one-shot pruning setup for all methods, so performance differences come from the pruning rule rather than recovery fine-tuning, extra data, or different evaluation conditions. Llama-3.1-8B is the primary model for the full suite, with Mistral-7B, Llama-2-7B, and Qwen2-7B used for cross-model validation and Llama-3.1-70B used for targeted scale validation \citep{touvron2023llama2,dubey2024llama3,jiang2023mistral,yang2024qwen2}. Unless stated otherwise, \(\LP\), \(\Conn\), and contribution correlations are estimated from 64 WikiText-2 training sequences of length 512 \citep{merity2017wikitext}; Appendix Table~\ref{tab:calibration} reports sensitivity to calibration size and source, including C4 \citep{raffel2020exploring}. For models without a main-pipeline HaloProtect run, the HaloProtect curve in Fig.~\ref{fig:cross-model-curves} is re-anchored from the side experiment by applying the HP/SCAR-LP ratio to the main SCAR-LP curve; absolute side-experiment values are reported in Appendix \S\ref{app:halo-extended}.

We evaluate WikiText-2 and C4 perplexity, zero-shot accuracy on HellaSwag, PIQA, BoolQ, WinoGrande, ARC-E, ARC-C, and OpenBookQA, and 5-shot MMLU accuracy \citep{zellers2019hellaswag,bisk2020piqa,clark2019boolq,sakaguchi2020winogrande,clark2018think,mihaylov2018openbookqa,hendrycks2021mmlu}. Baselines include channel-adapted Wanda, SparseGPT, activation-L2, OWL \citep{yin2024owl}, LLM-Pruner \citep{ma2023llmpruner}, FLAP \citep{an2024flap}, RIA \citep{zhang2024plug}, SlimLLM \citep{guo2025slimllm}, magnitude pruning, and random pruning. For methods originally designed for unstructured pruning, we aggregate per-weight saliencies over the grouped FFN channel: row \(i\) of \(W_{\text{up}}\), row \(i\) of \(W_{\text{gate}}\), and column \(i\) of \(W_{\text{down}}\). Thus, every method removes the same structured object under the same sparsity budget.

\ifdefined\NeurIPSSubmission
\begin{figure*}[!t]
    \centering
    \includegraphics[width=\textwidth]{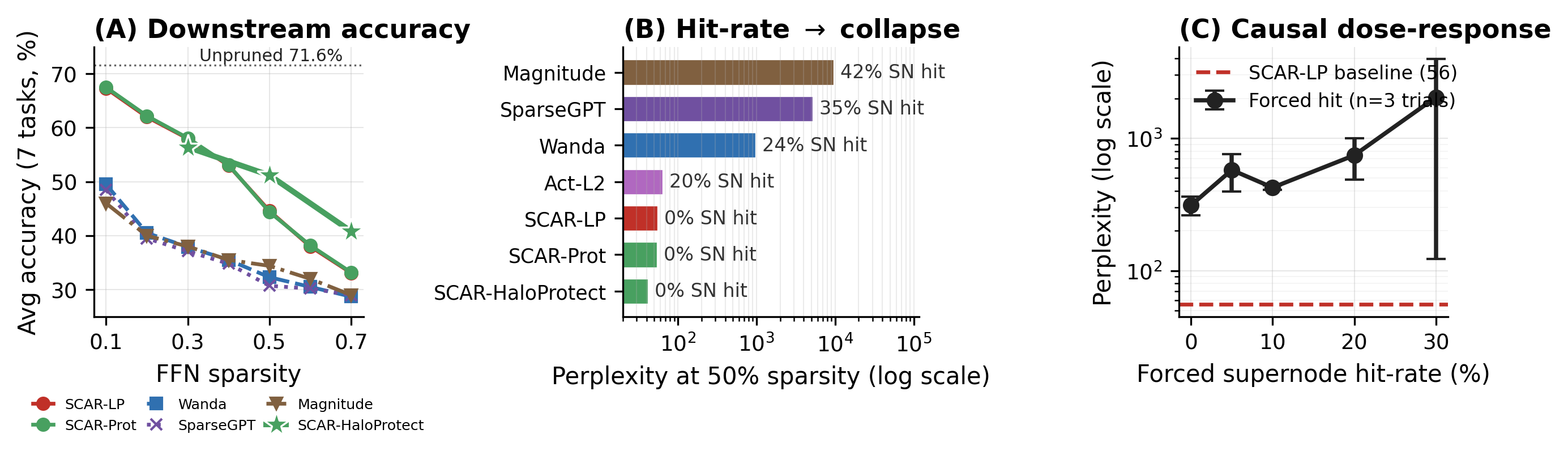}
    \caption{\textbf{Pruning probe and causal evidence on Llama-3.1-8B.}
    \textbf{(A)} SCAR variants retain higher downstream accuracy across sparsities.
    \textbf{(B)} At 50\% sparsity, supernode hit-rate tracks perplexity collapse.
    \textbf{(C)} Forced supernode removal gives a causal dose-response at fixed sparsity.}
    \label{fig:pruning-probe}
\end{figure*}
\fi

\paragraph{Main results.}
We start with the 8B comparison, where pruning quality and supernode hit-rate can be measured directly.

\ifdefined\NeurIPSSubmission
\begin{table}[!htbp]
\centering
\begin{minipage}[t]{0.66\textwidth}
\vspace{0pt}
\small
\setlength{\tabcolsep}{3.2pt}
\begin{tabular}{@{}lccc@{}}
\toprule
Method & PPL$\downarrow$ & Supernodes pruned (\%)$\downarrow$ & Avg. acc.$\uparrow$ \\
\midrule
Unpruned & 6.2 & 0.0 & 71.6 \\
Magnitude & 9790.2 & 41.7 & 34.4 \\
Wanda & 989.2 & 24.3 & 32.3 \\
SparseGPT & 5256.3 & 34.6 & 30.7 \\
Act. L2 & 65.5 & 19.6 & 43.3 \\
\midrule
SCAR-LP & 55.8 & \textbf{0.0} & 44.6 \\
SCAR-Prot & 54.8 & \textbf{0.0} & 44.4 \\
SCAR-Conn & 55.7 & \textbf{0.0} & 43.7 \\
SCAR-HaloProtect & \textbf{42.6} & \textbf{0.0} & \textbf{45.6} \\
\bottomrule
\end{tabular}

\end{minipage}
\hfill
\begin{minipage}[t]{0.30\textwidth}
\vspace{0pt}
\caption{\textbf{Llama-3.1-8B at 50\% FFN channel sparsity.} Average accuracy is over MMLU, HellaSwag, PIQA, BoolQ, WinoGrande, ARC-E, and ARC-C; OBQA and per-task rows appear in Appendix Table~\ref{tab:full-benchmarks}.}
\label{tab:main}
\end{minipage}
\end{table}
\else
\begin{table}[!b]
\centering
\caption{\textbf{Llama-3.1-8B at 50\% FFN channel sparsity.} We report each method's \emph{supernode hit-rate} alongside performance.}
\label{tab:main}
{\setlength{\tabcolsep}{2pt}\scriptsize
\resizebox{\columnwidth}{!}{
\begin{tabular}{@{}lcccccc@{}}
\toprule
Method & PPL$\downarrow$ & Supernodes pruned (\%)$\downarrow$ & MMLU & Hella & PIQA & BoolQ \\
\midrule
Unpruned & 6.2 & 0.0 & 69.0 & 60.0 & 84.0 & 80.0 \\
Magnitude (channel) & 9790.2 & 41.7 & \textbf{30.0} & 27.0 & 51.0 & 31.0 \\
Wanda (channel) & 989.2 & 24.3 & 21.0 & 26.0 & 56.0 & 30.0 \\
SparseGPT (channel) & 5256.3 & 34.6 & 22.0 & 22.0 & 56.0 & 32.0 \\
Act. L2 & 65.5 & 19.6 & 28.0 & 35.0 & 61.0 & \textbf{71.0} \\
SCAR-LP & 55.8 & \textbf{0.0} & 29.0 & 35.0 & \textbf{64.0} & 66.0 \\
SCAR-Prot & \textbf{54.8} & \textbf{0.0} & \textbf{30.0} & \textbf{37.0} & 63.0 & 66.0 \\
SCAR-Conn & 55.7 & \textbf{0.0} & 29.0 & 34.0 & \textbf{64.0} & 64.0 \\
\bottomrule
\end{tabular}
}
}
\end{table}
\fi

\ifdefined\NeurIPSSubmission
\else
\begin{figure*}[!t]
    \centering
    \includegraphics[width=0.88\textwidth]{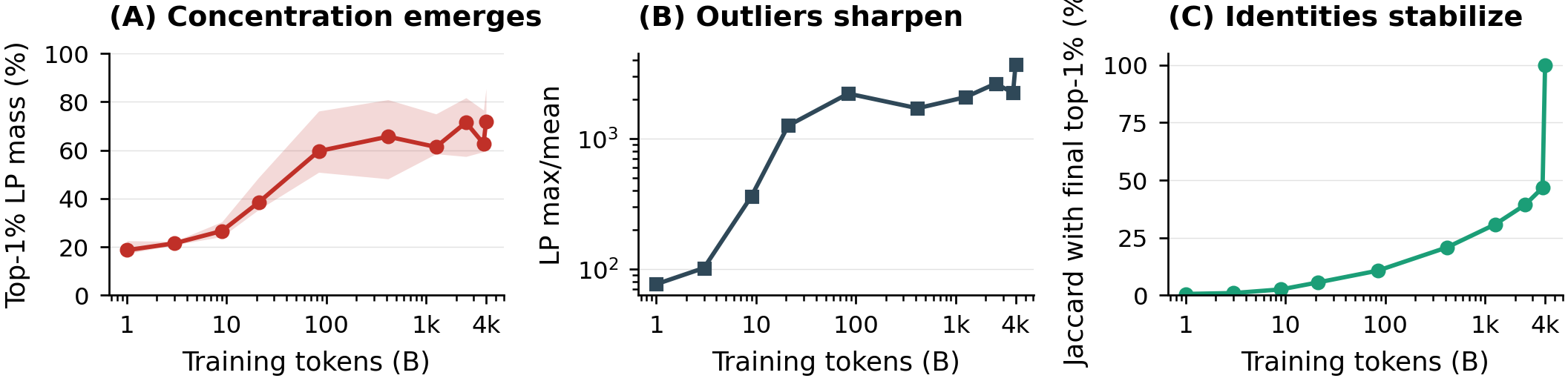}
    \caption{\textbf{OLMo-2 trajectory.} LP concentration, outlier strength, and final-set overlap increase through pretraining.}
    \label{fig:olmo-trajectory-main}
\end{figure*}
\fi

SCAR variants protect all supernodes and reach PPL between \PPLSCARLP{} and \PPLSCARProt{} at 50\% sparsity, compared with \PPLWandaChan{} for Wanda-channel and \PPLSparseGPTChan{} for SparseGPT-channel. The main-pipeline \SCAR{}-HaloProtect run reaches \PPLSCARHaloProtect{} PPL with average accuracy \AccSCARHaloProtect\%, improving the seven-task average while showing a single-task OBQA drop in Appendix Table~\ref{tab:full-benchmarks}. The side-pipeline HaloProtect sweeps in Appendix~\ref{app:halo-extended} come from a separate controlled $\tau$-sweep and should be interpreted through their HP/SCAR-LP ratios rather than compared by absolute PPL to Table~\ref{tab:main}.

The gap tracks supernode hit-rate: baselines prune $20$--$42\%$ of supernodes while SCAR variants prune $0\%$ (Fig.~\ref{fig:pruning-probe}B), and a forced-hit-rate sweep gives a clear dose--response (Fig.~\ref{fig:pruning-probe}C). At moderate sparsity, LP scoring alone already tends to place supernodes far above the pruning threshold, which explains why random protection can appear competitive in Table~\ref{tab:supernode-control}. The hard exclusion constraint is still useful because it makes the protected core explicit, prevents accidental core hits in variants and high-sparsity regimes, and separates LP-defined core protection from non-core ranking. Across the four models in Fig.~\ref{fig:cross-model-curves}, the SCAR-vs-baseline gap generally widens with sparsity; full per-task accuracies appear in Appendix Table~\ref{tab:full-benchmarks}.

\section{Cross-Model and Cross-Domain Consistency}
\label{sec:experiments}

The analysis so far has centered on Llama-3.1-8B. We next test how much of the pattern carries over to other models, larger scale, training checkpoints, and downstream tasks.

\paragraph{Cross-model validation.}
\label{sec:cross-model}

The same pruning pattern extends across Mistral-7B, Llama-2-7B, and Qwen2-7B (Appendix Table~\ref{tab:generalization}, Fig.~\ref{fig:cross-model-curves}), and LP concentration appears across these models as well (Appendix Table~\ref{tab:lp-concentration-models}). LP concentration also appears in OLMo-2-7B \citep{olmo2025olmo2}: at the final checkpoint, the top-1\% LP mass is \OLMoTopRhoMassFinalPct\%, higher than \TopRhoMassMedianPct\% on Llama-3.1-8B. The pruning advantage varies by model family, but the supernode concentration signature is broad; the HaloProtect rescue is largest in models and sparsity regimes where SCAR-LP itself begins to collapse (Fig.~\ref{fig:cross-model-curves}; Appendix \S\ref{app:halo-extended}).

\paragraph{Emergence during OLMo-2 pretraining.}
\label{sec:training-dynamics}

\ifdefined\NeurIPSSubmission
\begin{figure}[htbp]
    \centering
    \includegraphics[width=0.74\textwidth]{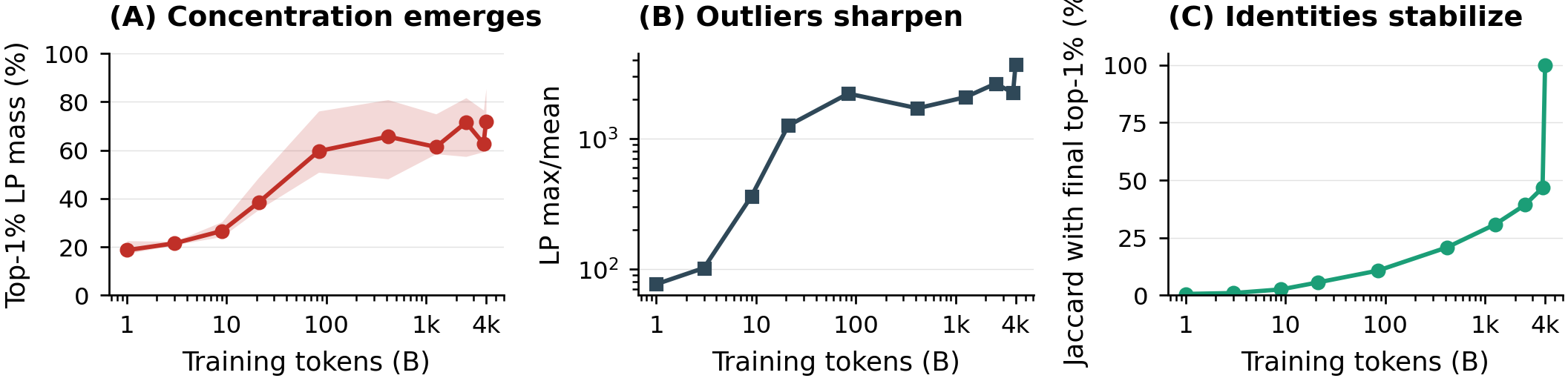}
    \caption{\textbf{OLMo-2 trajectory.} LP concentration, outlier strength, and final-set overlap increase through pretraining.}
    \label{fig:olmo-trajectory-main}
\end{figure}
\fi
OLMo-2-1124-7B's dense checkpoints let us track LP through pretraining (Fig.~\ref{fig:olmo-trajectory-main}): across $\sim$1B--$\sim$4T training tokens, top-1\% LP mass rises \OLMoTopRhoMassInitPct\%$\to$\OLMoTopRhoMassFinalPct\% and max/mean rises \OLMoMaxMeanInit{}$\to$\OLMoMaxMeanFinal{}. Jaccard overlap with the final top-1\% set rises from \OLMoJaccardInitFinalPct\% at the earliest checkpoint to \OLMoJaccardMidFinalPct\% at the last pre-final checkpoint; the final checkpoint is 100\% self-overlap by construction. The \emph{phenomenon} stabilizes earlier than the \emph{specific} channel identities; Appendix~\S\ref{app:olmo-training-dynamics} gives the OLMo pruning replication.

\paragraph{Targeted 70B validation.}
\label{sec:scale-70b}

On Llama-3.1-70B, both signatures persist. Under the matched factor-decomposition protocol, top-1\% exact-LP mass is 61.4\% at 70B versus 40.2\% at 8B; this is distinct from the per-layer median statistic of \TopRhoMassMedianPct\% reported above. SCAR-LP PPL is 10.71 at $s\!=\!50\%$ (Wanda 15.43, mag.\ 123.48, SparseGPT 224.41) and 36.29 at $s\!=\!70\%$ (Wanda 349.04). Additional 70B pruning curves and downstream results are in Fig.~\ref{fig:scale-70b-curves} and Tables~\ref{tab:scale-70b}/\ref{tab:scale-70b-benchmarks}; 8B calibration sensitivity is in Table~\ref{tab:calibration}.

\section{Conclusion, Limitations, and Impact}
\label{sec:conclusion}

We identify a small loss-critical core within LLM FFN layers. Under a Fisher-style activation--gradient proxy, channel sensitivity is highly concentrated, and direct core removal is far more damaging than matched non-core removal. In one-shot structured FFN pruning, baseline collapse is strongly associated with pruning LP-defined supernodes; LP-based scoring and explicit core protection avoid this failure, while read halos provide a secondary refinement at aggressive sparsity. The main limitations are calibration-dependent identities, FFN-only scope, strongest experimental coverage at 7--8B scale with targeted 70B/OLMo evidence, and the Fisher/Gauss--Newton approximation. Reliable pruning can reduce memory and compute, making LLMs cheaper to run and analyze while also lowering barriers to misuse; our contribution is methodological.

\ifdefined\ReleaseCodeURL
\paragraph{Reproducibility.}
Code for LP scoring, supernode/halo construction, SCAR pruning, and baselines is available at \ReleaseCodeURL. Derived artifacts, including numeric summaries, tables, figures, and release manifests, are available at \ReleaseArtifactsURL.
\fi

\ifdefined\NeurIPSSubmission
\bibliographystyle{plainnat}
\else
\bibliographystyle{icml2026}
\fi
\bibliography{references}

\newpage

\appendix

\setcounter{figure}{0}
\setcounter{table}{0}
\renewcommand{\thefigure}{A\arabic{figure}}
\renewcommand{\thetable}{A\arabic{table}}
\renewcommand{\theHfigure}{appendix.figure.\arabic{figure}}
\renewcommand{\theHtable}{appendix.table.\arabic{table}}

\section{Appendix Roadmap and Experimental Protocol}
\label{app:protocol}

The appendix is organized to follow the paper's evidence chain. Appendix~A gives the experimental protocol, default hyperparameters, runtime, and asset notes. Appendix~B gives the SCAR procedure, Appendix~C derives the loss proxy and redundancy score, Appendix~D collects the pruning tables and core-removal controls, Appendix~E gives additional mechanism and supernode-definition diagnostics, Appendix~F gives cross-model and scale results, Appendix~G covers OLMo training dynamics, and Appendices~H--I give the read-halo and HaloProtect details.

\paragraph{Models.}
Primary: Llama-3.1-8B (\texttt{meta-llama/Llama-3.1-8B}); additional validation on Mistral-7B, Llama-2-7B, and Qwen2-7B; targeted scale validation on Llama-3.1-70B. We evaluate in bf16 with a Hugging Face causal LM backend.

\paragraph{Calibration (metric estimation).}
Unless stated otherwise, we estimate \(\LP\), \(\Conn\), and redundancy statistics from the WikiText-2 training split using 64 sequences of length 512 (next-token prediction loss). Calibration examples are tokenized with the model tokenizer and sampled as deterministic contiguous token blocks with a fixed base seed; the C4, code, and arXiv calibration variants use the same sequence length and sampling convention on their respective calibration splits.

\paragraph{Structured channel pruning protocol.}
We prune FFN \emph{channels} (SwiGLU intermediate coordinates): removing channel \(i\) deletes row \(i\) from \(W_{\mathrm{gate}}, W_{\mathrm{up}}\) and column \(i\) from \(W_{\mathrm{down}}\). We target a global FFN channel sparsity \(s\) with per-layer caps \(s_\ell^{\max}=70\%\) (unless otherwise specified) and evaluate all methods in a one-shot (no fine-tuning) setting.
For channel-adapted versions of weight-level baselines, let
\[
G_i=\{W_{\mathrm{up}}[i,:],\, W_{\mathrm{gate}}[i,:],\, W_{\mathrm{down}}[:,i]\}
\]
be the parameter group associated with FFN channel \(i\). If a baseline provides per-weight saliencies \(S_p\), we use
\[
S_i^{\mathrm{channel}}=\sum_{p\in G_i} S_p .
\]
All groups have the same size within a model, so sum and mean aggregation induce the same ranking.

\paragraph{Perplexity.}
WikiText-2 perplexity follows a standard blockwise protocol: concatenate the test split and evaluate contiguous 2048-token blocks (no padding). We also report C4 validation perplexity where indicated.

\paragraph{Downstream tasks.}
We evaluate zero-/few-shot accuracy on MMLU (5-shot), HellaSwag, PIQA, BoolQ, WinoGrande, ARC-E, ARC-C, and OpenBookQA with fixed prompt settings, using the \texttt{lm-evaluation-harness} of \citet{gao2023evalharness}.

\paragraph{Randomness and seeds.}
Where stochasticity is present (e.g., random masks), we report means over multiple trials with a fixed base seed and deterministic per-trial seeding.

\paragraph{Default hyperparameters.}
\label{app:hyperparameters}
Table~\ref{tab:hyperparameters} lists the default supernode, halo, calibration, and pruning settings. These values define the paper's default configuration but are not meant to be universal constants; the diagnostics remain stable under moderate changes around them.

\begin{table}[H]
\centering
\caption{\textbf{Default hyperparameters.}}
\label{tab:hyperparameters}
\small
\begin{tabular}{@{}lp{0.62\columnwidth}c@{}}
\toprule
Parameter & Description & Value \\
\midrule
$\rho$ & Supernode fraction & 1\% \\
$\eta$ & Halo fraction & 10\% \\
$\eta_r$ & Read-halo fraction & 10\% \\
$\tau$ & HaloProtect fraction within read-halo set & 10\% \\
$K$ & Conn top-$K$ hidden dims & 256 \\
$\gamma$ & Protection rank power & 8 \\
$\alpha$ & Protection floor & 0.2 \\
Seq.\ length & Calibration sequence length (metric estimation) & 512 \\
$n_{\text{seq}}$ & Calibration sequences & 64 \\
$s_\ell^{\max}$ & Max per-layer sparsity & 70\% \\
\bottomrule
\end{tabular}
\end{table}

\paragraph{Runtime.}
\label{app:runtime}
Table~\ref{tab:runtime} breaks SCAR into its main timed stages on a single A100 GPU. The backward pass for LP estimation is the main extra cost relative to activation-only pruning.

\begin{table}[H]
\centering
\caption{\textbf{Runtime breakdown} for Llama-3.1-8B on a single A100. The forward-only row is a baseline comparison and is not included in the SCAR-specific subtotal.}
\label{tab:runtime}
\small
\begin{tabular}{@{}lc@{}}
\toprule
Phase & Time \\
\midrule
Forward-only activation collection baseline & 8 min \\
Forward + backward LP collection/accumulation & 15 min \\
Halo/connectivity computation & 5 min \\
Directed redundancy & 12 min \\
Pruning \& evaluation & 3 min \\
\midrule
\textbf{SCAR-specific subtotal} & \textbf{35 min} \\
\bottomrule
\end{tabular}
\end{table}

SCAR's additional cost relative to magnitude or activation-only pruning is the backward pass needed for gradient signals \(s_i\). In our implementation, LP estimation on Llama-3.1-8B takes about 15 minutes on a single A100, compared with about 8 minutes for forward-only activation collection; weight-magnitude pruning requires no calibration data. The full SCAR-Prot variant, including connectivity and directed redundancy, sums to about 35 minutes before ordinary setup and job-launch overhead.

\paragraph{External assets.}
We use publicly released model families and benchmarks under their provider terms or published licenses: Llama-2 and Llama-3.1 under Meta's Llama licenses, Mistral-7B under its released model license, Qwen2 and OLMo-2 under their released model terms, and WikiText-2, C4, MMLU, HellaSwag, PIQA, BoolQ, WinoGrande, ARC, and OpenBookQA under their published dataset licenses or access terms. We do not redistribute model weights or benchmark data.
\begin{table}[H]
\centering
\caption{\textbf{External assets and access terms.} We use publicly released assets through their official model or dataset cards and do not redistribute weights or benchmark data.}
\label{tab:asset-licenses}
\small
\begin{tabular}{@{}p{0.30\columnwidth}p{0.62\columnwidth}@{}}
\toprule
Asset & License or access terms used \\
\midrule
Llama-2, Llama-3.1 & Meta Llama community licenses / model access terms \\
Mistral-7B & Released Mistral model license / model-card terms \\
Qwen2-7B & Released Qwen2 model license / model-card terms \\
OLMo-2-7B & Released OLMo model license / model-card terms \\
WikiText-2, C4 & Published dataset licenses and dataset-card terms \\
MMLU, HellaSwag, PIQA, BoolQ, WinoGrande, ARC, OpenBookQA & Published benchmark licenses and dataset-card terms \\
\bottomrule
\end{tabular}
\end{table}

\section{SCAR Algorithm and Scores}
\label{app:algorithm}

This section gives the full procedural version of SCAR. The algorithm separates metric estimation, constrained channel selection, and structured channel removal, following the three steps used in the main text.

\begin{algorithm}[H]
\caption{\SCAR{}: Supernode-Constrained Structured FFN Channel Pruning}
\label{alg:scar}
\begin{algorithmic}[1]
\Require model; calibration tokens $\mathcal{D}$; target global sparsity $s$
\Require supernode fraction $\rho$, write-halo fraction $\eta$, read-halo fraction $\eta_r$, HaloProtect fraction $\tau$, per-layer caps $s_\ell^{\max}$
\State \textbf{Phase 1: Metric estimation (per layer)}
\For{each FFN layer $\ell$}
    \State Accumulate $\LP_i = \frac{1}{2}\mathbb{E}[(u_i s_i)^2]$ for all channels $i$
    \State Select supernodes $\mathcal{M}_\ell \leftarrow$ top-$\rho$ channels by $\LP_i$
    \State Compute supernode write pattern $a = \sum_{s \in \mathcal{M}_\ell} |v_s|$
    \State Define support $\mathcal{S}_\ell \leftarrow \mathrm{TopK}(a)$
    \State Compute connectivity $\Conn_j$ for all non-supernodes $j$
    \State Halo $\mathcal{H}_\ell \leftarrow$ top-$\eta$ non-supernodes by $\Conn_j$
    \State For each $j \in \mathcal{H}_\ell$, compute $\Red^{\rightarrow \mathcal{M}}_j$
    \State Convert to $\Protect_j$; set $\Protect_j \leftarrow 1$ for $j \notin \mathcal{H}_\ell$
    \State If $\ell < L-1$, compute $\mathrm{ReadConn}_j$ in layer $\ell+1$ using $\mathcal{S}_\ell$
    \State If $\ell < L-1$, set read halo $\mathcal{R}_{\ell+1} \leftarrow$ top-$\eta_r$ channels by $\mathrm{ReadConn}$
    \State If $\ell < L-1$, set HaloProtect set $\mathcal{B}_{\ell+1} \leftarrow$ top-$\tau$ fraction of $\mathcal{R}_{\ell+1}$ by $\mathrm{ReadConn}$
\EndFor
\State Set undefined boundary HaloProtect sets (e.g., $\mathcal{B}_0$) to $\emptyset$
\State \textbf{Phase 2: Selection under constraints}
\State Initialize prune set $\mathcal{P} \leftarrow \emptyset$
\For{each layer $\ell$}
    \State Compute scores for the selected variant:
    \State \hspace{1em}\SCAR-LP: Score$_i \leftarrow \LP_i$
    \State \hspace{1em}\SCAR-Prot: Score$_i \leftarrow \LP_i \cdot \Protect_i$
    \State \hspace{1em}\SCAR-Conn: Score$_i \leftarrow \LP_i \cdot [(1-\Conn_i)+\Conn_i\Protect_i]$
    \State \hspace{1em}\SCAR{}-HaloProtect: Score$_i \leftarrow \LP_i$
\EndFor
\State Build a global candidate pool of all non-supernode channels with layer IDs and scores
\State For \SCAR{}-HaloProtect, additionally remove channels in $\mathcal{B}_\ell$ from this pool
\State Sort candidates by increasing score
\For{candidate channel $i$ in sorted order}
    \State If global sparsity target $s$ is reached, stop
    \State If adding $i$ does not exceed its layer cap $s_\ell^{\max}$, add $i$ to $\mathcal{P}$
\EndFor
\State \textbf{Phase 3: Structured removal}
\State Remove pruned channels from $W_{\text{up}}, W_{\text{gate}}, W_{\text{down}}$
\State \textbf{return} pruned model
\end{algorithmic}
\end{algorithm}

\subsection{Protection and Connectivity Scores}
\label{app:protection-score}

For halo channel $j$, the protection score is:
\begin{equation}
    \Protect_j = \alpha + (1 - \alpha)(1 - r_j^\gamma),
\end{equation}
where
\[
    r_j =
    \frac{\operatorname{rank}_{\mathrm{asc}}(\Red^{\rightarrow \mathcal{M}}_j)-1}{|\mathcal{H}|-1}
\]
is the zero-indexed ascending rank of \(j\)'s redundancy within the halo.
Default: $\alpha = 0.2$, $\gamma = 8$.

The \SCAR{}-LP, \SCAR{}-Prot, and \SCAR{}-Conn variants use the following channel scores for non-supernode channels:
\begin{align}
    \mathrm{Score}_i^{\mathrm{LP}} &= \LP_i, \\
    \mathrm{Score}_i^{\mathrm{Prot}} &= \LP_i \Protect_i, \\
    \mathrm{Score}_i^{\mathrm{Conn}} &= \LP_i\big[(1-\Conn_i)+\Conn_i\Protect_i\big],
\end{align}
where $\Conn_i \in [0,1]$ is the write-connectivity score $\Conn_j = \|v_j[\mathcal{S}]\|_1 / \|v_j\|_1$ defined in \S\ref{sec:halos}. Supernodes are excluded from the candidate prune set for all variants; \SCAR{}-HaloProtect adds the separate hard read-halo exclusion in Algorithm~\ref{alg:scar}.

\paragraph{Intuition.}
Because lower scores are pruned first, multiplying LP by a smaller protection value makes highly redundant halo channels more likely to be removed.
Channels with low redundancy (low $r_j$) get $\Protect_j \approx 1$ and retain their LP score.
Channels with high redundancy, corresponding to high \(r_j\), get \(\Protect_j \approx \alpha\) and become easier to prune.
The floor $\alpha > 0$ prevents over-penalizing noisy redundancy estimates.

\section{Derivations}
\label{app:derivations}

This section provides the derivations behind the two main quantities used in the paper: the Fisher-style channel loss proxy and the Gaussian-inspired redundancy score. These derivations clarify the approximations and show where the resulting scores enter the pruning pipeline.

\subsection{Loss Proxy Derivation}

Our loss proxy builds on the classical second-order sensitivity framework introduced by \citet{lecun1990optimal} (Optimal Brain Damage), extended by \citet{hassibi1993obs} (Optimal Brain Surgeon) to off-diagonal Hessian terms, and brought to modern neural networks by \citet{molchanov2019importance}; \citet{kurtic2022obert} and \citet{frantar2021mfac} adapt this framework to large transformers via efficient empirical-Fisher approximations.

\begin{figure}[H]
    \centering
    \includegraphics[width=\columnwidth]{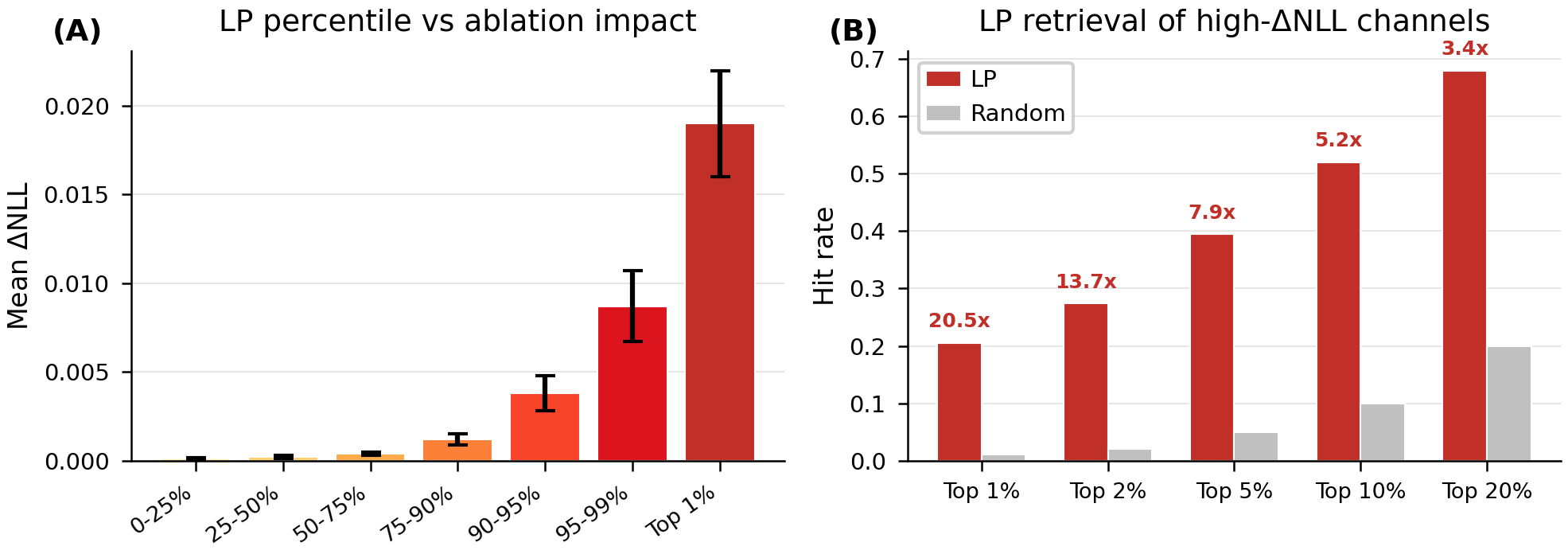}
    \caption{\textbf{Loss proxy validation.} Higher LP percentiles correspond to larger ablation effects.}
    \label{fig:lp-proxy-diagnostics}
\end{figure}

Masking channel $i$ induces $\Delta u = -u_i e_i$ and $\Delta y = -u_i v_i$.
Under Fisher/Gauss--Newton approximation, the expected loss change is:
\begin{align}
    \mathbb{E}[\Delta \mathcal{L}] &\approx \frac{1}{2} \mathbb{E}[\Delta y^\top F_y \Delta y] \\
    &= \frac{1}{2} \mathbb{E}[u_i^2 \cdot v_i^\top F_y v_i] \\
    &\approx \frac{1}{2} \mathbb{E}[u_i^2 \cdot (v_i^\top g_y)^2] \\
    &= \frac{1}{2} \mathbb{E}[(u_i s_i)^2] = \LP_i.
\end{align}
The key approximation replaces the Fisher information $F_y$ with its empirical estimate $g_y g_y^\top$, which is standard in scalable importance estimation \citep{molchanov2019importance}.
We use this as a ranking proxy and omit the first-order term, as in empirical-Fisher pruning approximations; Appendix Fig.~\ref{fig:lp-proxy-diagnostics} validates the top-tail ranking against direct ablation.
The proxy is deliberately local: it scores the effect of masking one channel through its activation--gradient contribution and does not attempt to model all off-diagonal interactions among channels in a joint pruning mask. Those interactions are instead tested empirically through core-removal controls and the structured pruning experiments.

\paragraph{Relation to weight-level pruning.}
Standard unstructured pruning methods like \citet{han2015learning} use magnitude-based criteria.
Our loss proxy extends this to structured channel removal by jointly considering activation magnitude ($u_i$) and gradient coupling ($s_i$), capturing loss-sensitive importance beyond pure magnitude.

\subsection{Gaussian MI Motivation for Redundancy}

For loss-relevant contributions \(q_i = u_i \cdot s_i\) and \(q_j = u_j \cdot s_j\), the exact Gaussian mutual information provides a useful reference point.
For jointly Gaussian \((q_i, q_j)\) with Pearson correlation \(\rho_{ij}\):
\begin{equation}
    I(q_i; q_j) = -\frac{1}{2} \log(1 - \rho_{ij}^2).
\end{equation}
This is the analytic MI for a bivariate Gaussian, which depends only on the correlation coefficient.
Our main redundancy score is a positive-only variant motivated by this expression:
\begin{align}
    \rho^+_{ij} &= \max(\rho_{ij},0), \\
    \Red(i,j) &= -\frac{1}{2}\log\left(1-(\rho^+_{ij})^2\right).
\end{align}
It measures how much positively correlated information about \(q_i\)'s contribution to loss is already captured by \(q_j\), without treating anti-correlations as substitutes.
We do not assume that LLM channel contributions are actually Gaussian; the formula is used as a computationally cheap, monotone correlation-based redundancy proxy. The relevant claims depend on the induced ranking and are checked through conditional ablations and pruning controls rather than on calibrated mutual-information values.
Channels with high directed redundancy to the supernode core are candidates for lower-risk removal because their loss-relevant contributions may be partially predictable from protected supernodes.
For the directed score used by \SCAR{}-Prot and \SCAR{}-Conn, we aggregate pairwise redundancy over the strongest supernode neighbors:
\[
    \Red^{\rightarrow \mathcal{M}}_j =
    \frac{1}{k}\sum_{s \in \mathrm{Top}\text{-}k(\mathcal{M};j)} \Red(j,s),
\]
where \(\mathrm{Top}\text{-}k(\mathcal{M};j)\) returns the \(k\) supernodes with largest \(\Red(j,s)\) for channel \(j\). We use \(k=5\) by default.

\clearpage
\section{Pruning Controls and Main Result Tables}
\label{app:extended}

This section collects the tables needed to audit the main pruning claims: which supernodes each method removes, full task-level results at 50\% sparsity, direct core-removal controls, calibration sensitivity, and paper-faithful unstructured baselines. Mechanistic diagnostics and cross-model results are separated into the following appendices.

\begin{table}[H]
\centering
\caption{\textbf{Supernode hit-rate at \MainSparsityPct\% sparsity.} Fraction of supernodes pruned by each method in the main comparison.}
\label{tab:supernode-hit-rate}
\small
\begin{tabular}{@{}lcc@{}}
\toprule
Method & Supernodes pruned (\%)$\downarrow$ & PPL$\downarrow$ \\
\midrule
Magnitude (channel) & 41.7 (1910/4576) & 9790.2 \\
Act-L2 (channel) & 19.6 (897/4576) & 65.5 \\
Wanda (channel) & 24.3 (1112/4576) & 989.2 \\
SparseGPT (channel) & 34.6 (1584/4576) & 5256.3 \\
OWL (channel) & 22.0 (1009/4576) & 585.4 \\
LLM-Pruner (channel) & 26.8 (1228/4576) & 607.3 \\
FLAP (channel) & 59.4 (2719/4576) & 220167.1 \\
RIA (channel) & 15.5 (710/4576) & 109.7 \\
SlimLLM (channel) & 22.0 (1005/4576) & 178.2 \\
SCAR-LP & \textbf{0.0 (0/4576)} & 55.8 \\
SCAR-Prot & \textbf{0.0 (0/4576)} & 54.8 \\
SCAR-Conn & \textbf{0.0 (0/4576)} & 55.7 \\
SCAR-HaloProtect & \textbf{0.0 (0/4576)} & 42.6 \\
\bottomrule
\end{tabular}

\end{table}

\begin{table*}[!htbp]
\centering
\caption{\textbf{Complete benchmark results} for Llama-3.1-8B at 50\% FFN channel sparsity, including the separate main-pipeline HaloProtect run.}
\label{tab:full-benchmarks}
\small
\ifdefined\NeurIPSSubmission
\resizebox{\textwidth}{!}{
\begin{tabular}{@{}lccccccccc@{}}
\toprule
Method & PPL$\downarrow$ & MMLU & Hella & PIQA & BoolQ & WinoG & ARC-E & ARC-C & OBQA \\
\midrule
Unpruned & 6.2 & 69.0 & 60.0 & 84.0 & 80.0 & 67.0 & 85.0 & 56.0 & 76.0 \\
Random & 17428.5 & 16.0 & 31.0 & 46.0 & 30.0 & 51.0 & 26.0 & 24.0 & 21.0 \\
Magnitude (channel) & 9790.2 & \textbf{30.0} & 27.0 & 51.0 & 31.0 & 52.0 & 28.0 & 22.0 & 32.0 \\
Wanda (channel) & 989.2 & 21.0 & 26.0 & 56.0 & 30.0 & 43.0 & 27.0 & 25.0 & 29.0 \\
SparseGPT (channel) & 5256.3 & 22.0 & 22.0 & 56.0 & 32.0 & 53.0 & 26.0 & 24.0 & 28.0 \\
OWL (channel) & 585.4 & 21.0 & 28.0 & 58.0 & 35.0 & 47.0 & 28.0 & \textbf{27.0} & 29.0 \\
LLM-Pruner (channel) & 607.3 & 23.0 & 21.0 & 44.0 & 30.0 & 47.0 & 30.0 & 21.0 & 31.0 \\
FLAP (channel) & 220167.1 & 28.0 & 20.0 & 56.0 & 64.0 & 51.0 & 29.0 & 19.0 & 33.0 \\
RIA (channel) & 109.7 & 27.0 & 33.0 & 64.0 & 68.0 & 53.0 & 34.0 & 22.0 & 27.0 \\
SlimLLM (channel) & 178.2 & 22.0 & 32.0 & \textbf{67.0} & 58.0 & 47.0 & 30.0 & 23.0 & 22.0 \\
Act. L2 & 65.5 & 28.0 & 35.0 & 61.0 & \textbf{71.0} & 50.0 & \textbf{44.0} & 23.0 & 29.0 \\
SCAR-LP & 55.8 & 29.0 & 35.0 & 64.0 & 66.0 & 55.0 & 37.0 & 26.0 & \textbf{34.0} \\
SCAR-Prot & 54.8 & \textbf{30.0} & 37.0 & 63.0 & 66.0 & \textbf{57.0} & 35.0 & 23.0 & \textbf{34.0} \\
SCAR-Conn & 55.7 & 29.0 & 34.0 & 64.0 & 64.0 & 54.0 & 37.0 & 24.0 & \textbf{34.0} \\
SCAR-HaloProtect & \textbf{42.6} & 28.0 & \textbf{38.0} & 65.0 & 69.0 & \textbf{57.0} & 36.0 & 26.0 & 21.0 \\
\bottomrule
\end{tabular}
}
\else
\begin{tabular}{@{}lccccccccc@{}}
\toprule
Method & PPL$\downarrow$ & MMLU & Hella & PIQA & BoolQ & WinoG & ARC-E & ARC-C & OBQA \\
\midrule
Unpruned & 6.2 & 69.0 & 60.0 & 84.0 & 80.0 & 67.0 & 85.0 & 56.0 & 76.0 \\
Random & 17428.5 & 16.0 & 31.0 & 46.0 & 30.0 & 51.0 & 26.0 & 24.0 & 21.0 \\
Magnitude (channel) & 9790.2 & \textbf{30.0} & 27.0 & 51.0 & 31.0 & 52.0 & 28.0 & 22.0 & 32.0 \\
Wanda (channel) & 989.2 & 21.0 & 26.0 & 56.0 & 30.0 & 43.0 & 27.0 & 25.0 & 29.0 \\
SparseGPT (channel) & 5256.3 & 22.0 & 22.0 & 56.0 & 32.0 & 53.0 & 26.0 & 24.0 & 28.0 \\
OWL (channel) & 585.4 & 21.0 & 28.0 & 58.0 & 35.0 & 47.0 & 28.0 & \textbf{27.0} & 29.0 \\
LLM-Pruner (channel) & 607.3 & 23.0 & 21.0 & 44.0 & 30.0 & 47.0 & 30.0 & 21.0 & 31.0 \\
FLAP (channel) & 220167.1 & 28.0 & 20.0 & 56.0 & 64.0 & 51.0 & 29.0 & 19.0 & 33.0 \\
RIA (channel) & 109.7 & 27.0 & 33.0 & 64.0 & 68.0 & 53.0 & 34.0 & 22.0 & 27.0 \\
SlimLLM (channel) & 178.2 & 22.0 & 32.0 & \textbf{67.0} & 58.0 & 47.0 & 30.0 & 23.0 & 22.0 \\
Act. L2 & 65.5 & 28.0 & 35.0 & 61.0 & \textbf{71.0} & 50.0 & \textbf{44.0} & 23.0 & 29.0 \\
SCAR-LP & 55.8 & 29.0 & 35.0 & 64.0 & 66.0 & 55.0 & 37.0 & 26.0 & \textbf{34.0} \\
SCAR-Prot & 54.8 & \textbf{30.0} & 37.0 & 63.0 & 66.0 & \textbf{57.0} & 35.0 & 23.0 & \textbf{34.0} \\
SCAR-Conn & 55.7 & 29.0 & 34.0 & 64.0 & 64.0 & 54.0 & 37.0 & 24.0 & \textbf{34.0} \\
SCAR-HaloProtect & \textbf{42.6} & 28.0 & \textbf{38.0} & 65.0 & 69.0 & \textbf{57.0} & 36.0 & 26.0 & 21.0 \\
\bottomrule
\end{tabular}

\fi
\end{table*}

\paragraph{Direct core-removal and calibration controls.}
The first control below forces or avoids direct hits to the LP-defined core, isolating supernode removal as the manipulated variable. The second checks that the 50\% SCAR-Conn result is not an artifact of one calibration source or sample count.

\begin{table}[H]
\centering
\caption{\textbf{Supernode ablation control.} Removing supernodes first is catastrophic.}
\label{tab:supernode-control}
\small
\begin{tabular}{@{}p{0.44\columnwidth}cp{0.23\columnwidth}@{}}
\toprule
Strategy & PPL & Relative \\
\midrule
Remove supernodes early & 811894.1 & catastrophic \\
Wanda (channel) & 989.2 & poor \\
Protect LP supernodes (ablation) & 55.8 & core protected \\
Protect random channels (ablation) & 55.9 $\pm$ 0.5 & similar at 50\% \\
SCAR-Prot (protect supernodes) & \textbf{54.8} & SCAR variant \\
SCAR-Conn (protect supernodes) & 55.7 & SCAR variant \\
\bottomrule
\end{tabular}

\end{table}

\begin{table}[H]
\centering
\caption{\textbf{Calibration-source sensitivity.} SCAR-Conn on Llama-3.1-8B at 50\% FFN sparsity, evaluated by WikiText-2 perplexity under different calibration sources and sample counts.}
\label{tab:calibration}
\small
\begin{tabular}{@{}llc@{}}
\toprule
Dataset & \# seqs & PPL \\
\midrule
WikiText-2 & 128 & 48.4 \\
WikiText-2 & 64 & 55.7 \\
WikiText-2 & 32 & 79.3 \\
C4 & 128 & 40.3 \\
Mixed (Wiki + C4) & 128 & 49.9 \\
\bottomrule
\end{tabular}

\end{table}

\subsection{Paper-Faithful Unstructured Baselines}
\label{app:unstructured-baselines}

Many popular LLM pruning methods are designed for \emph{unstructured} weight sparsity.
For completeness, we also report paper-faithful \emph{unstructured} Wanda and SparseGPT baselines, which prune individual FFN weights (FFN-only scope) using the original scoring/reconstruction rules.
These appendix results answer a different question from the main text: they show how the original methods behave in their native unstructured setting, whereas our main results study the harder and more deployment-relevant regime of grouped FFN channel removal.

\begin{table}[H]
\centering
\caption{\textbf{Unstructured baseline results} (FFN-only scope) for Llama-3.1-8B at 50\% sparsity.}
\label{tab:unstructured-baselines}
\small
\begin{tabular}{@{}p{0.72\columnwidth}c@{}}
\toprule
Method & PPL$\downarrow$ \\
\midrule
Unpruned & 6.2 \\
Wanda (unstructured) & 9.3 \\
SparseGPT (unstructured) & \textbf{8.9} \\
\bottomrule
\end{tabular}

\end{table}

\clearpage
\section{Mechanistic Diagnostics and Supernode Definitions}

This appendix gathers supporting evidence that is not already shown as a main figure: LP is not reducible to simple activation or norm scores, supernodes cannot be replaced by mean activations, read/write halo variants behave differently, and LP-defined supernodes differ from activation-defined outliers. The main text already contains the supernode dose--response and conditional halo-ablation panels in Figs.~\ref{fig:pruning-probe}C and \ref{fig:halo-structure}C.

\subsection{Mechanism Controls Summary}

The following controls test whether LP concentration can be reduced to simpler magnitude-based explanations. They show that activation power and weight norms explain part of the structure, but not the full loss-sensitive ranking.

\begin{table}[H]
\centering
\caption{\textbf{Mechanism controls summary.} Additional diagnostics showing that LP is not explained by activation power or weight norms alone. LP-vs-ablation validation is shown separately in Fig.~\ref{fig:lp-proxy-diagnostics}; the main dose-response and conditional halo-ablation controls are shown in Figs.~\ref{fig:pruning-probe}C and \ref{fig:halo-structure}C.}
\label{tab:mechanism-controls}
\small
\begin{tabular}{@{}p{0.72\columnwidth}c@{}}
\toprule
Mechanism control & Value \\
\midrule
Spearman $\rho$(log LP, log ActPower) (median over layers) & 0.321 \\
Spearman $\rho$(log LP, log $\|v_i\|$) (median over layers) & -0.003 \\
Spearman $\rho$(log LP, log $\|W_{up}[i]\|$) (median over layers) & -0.005 \\
Spearman $\rho$(log LP, log $\|W_{gate}[i]\|$) (median over layers) & 0.154 \\
\bottomrule
\end{tabular}

\end{table}

\subsection{Additional Mechanism Figures}

The main text focuses on the strongest mechanistic evidence. We therefore keep only appendix-only diagnostics here: LP validation, LP-versus-activation extensions, and the mean-replacement control. The read-halo evidence is consolidated in Appendices~\ref{app:halo-rescue}--\ref{app:halo-extended}.

\paragraph{Loss proxy validation.}
Single-channel ablation effects are sparse and noisy, so global rank correlation over all channels is not the most informative validation target. The relevant diagnostic is top-tail enrichment: high-LP percentiles have substantially larger mean ablation loss and retrieve high-\(\Delta\)NLL channels more often than activation-power baselines.

\paragraph{LP vs.\ activation supernodes.}
Figure~\ref{fig:halo-structure}D shows the central 8B overlap result: LP-defined and activation-defined top-1\% channel sets overlap only weakly, so activation outliers are not simply recovering the same supernode identities.
Appendix \S\ref{app:lp-vs-act-supernodes} extends this comparison with 70B matched-control pruning, factor-decomposition results, and cross-domain stability diagnostics.

\begin{figure}[H]
    \centering
    \includegraphics[width=0.68\columnwidth]{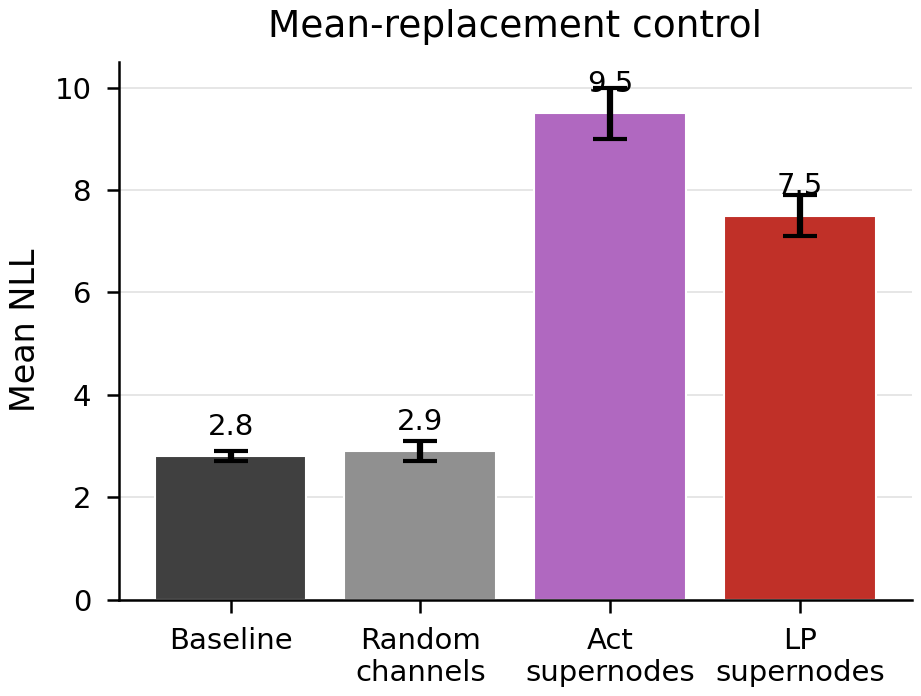}
    \caption{\textbf{Mean-replacement control.} Replacing supernode activations with their mean is damaging; random channels are unaffected.}
    \label{fig:mean-replacement-control}
\end{figure}

\paragraph{Mean-replacement control.}
We test whether supernodes can be trivially replaced by their mean activation. Figure~\ref{fig:mean-replacement-control} shows that this intervention is catastrophic for supernodes but negligible for random channels.

\subsection{LP-Defined vs Activation-Defined Supernodes}
\label{app:lp-vs-act-supernodes}

\begin{figure}[H]
    \centering
    \includegraphics[width=\columnwidth]{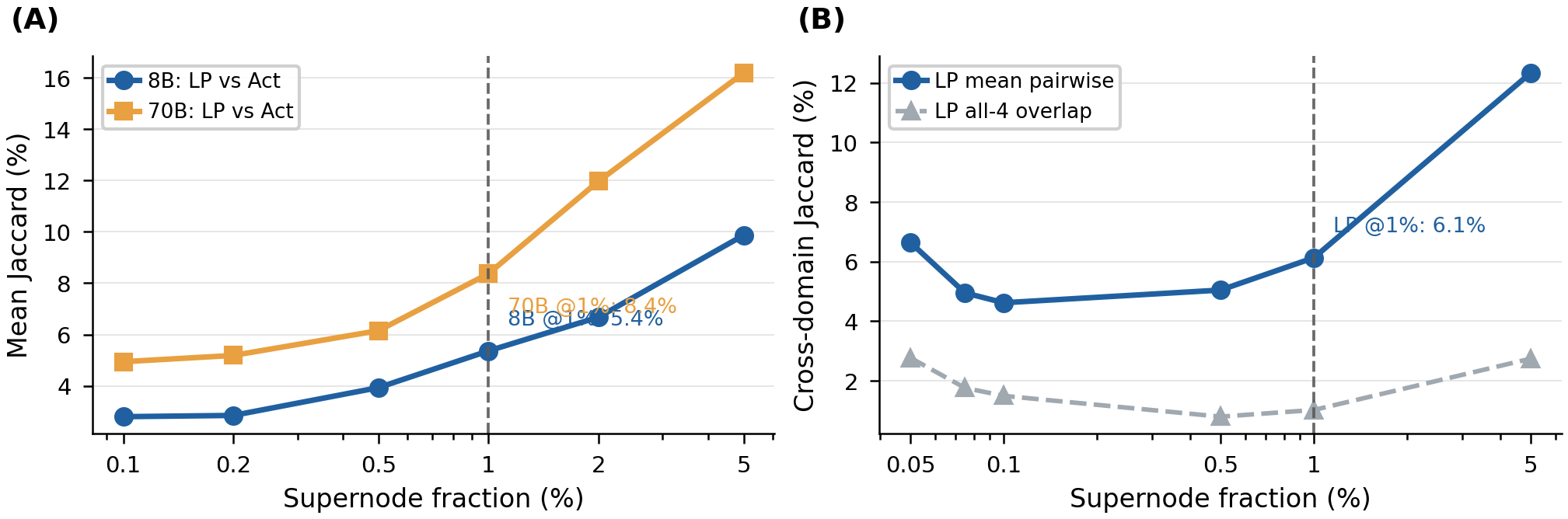}
    \caption{\textbf{Threshold sensitivity of LP-vs-activation overlap and LP domain stability.} \textbf{(A)} Mean LP-vs-activation Jaccard rises with the supernode fraction: on 8B it increases from 2.8\% at 0.1\% to 5.4\% at 1\% and 9.9\% at 5\%; on 70B it increases from 4.9\% at 0.1\% to 8.4\% at 1\% and 16.2\% at 5\%. \textbf{(B)} LP cross-domain stability is also fraction-sensitive: mean pairwise Jaccard across WikiText/C4/Code/arXiv rises from 4.6\% at 0.1\% to 6.1\% at 1\% and 12.3\% at 5\%. The default 1\% setting therefore lies in a low-overlap regime, but the trend with fraction is smooth rather than discontinuous.}
    \label{fig:lp-vs-activation-overlap}
\end{figure}

A natural question is whether LP-defined supernodes differ from activation-defined ones, or whether they are mostly the same channels under another name. The main text already shows the core 8B overlap result in Fig.~\ref{fig:halo-structure}D; here we add the matched 70B control and the scale and stability analyses that help interpret it.
We find that the two definitions identify substantially different channel sets at the default 1\% threshold (Jaccard $\approx$5.4\% at 8B, $\approx$8.4\% at 70B; Spearman correlation $\approx$0.26 at 8B, $\approx$0.31 at 70B).
This overlap is itself threshold-sensitive: when the protected core is made smaller, the LP--activation intersection drops further (e.g., to $\approx$2.8\% at 0.1\% on 8B), while larger fractions produce gradually higher overlap (Fig.~\ref{fig:lp-vs-activation-overlap}).
This is the sensitivity question most relevant to the paper's main claims, because the supernode fraction determines which channels are treated as the protected core, whereas choices such as the connectivity top-$K$ only affect the secondary halo construction.
Despite this low overlap, when the \emph{pruning score} is held fixed to the SCAR-LP score and only the \emph{protected set} definition is changed, pruning outcomes are very similar: on Llama-3.1-70B at 50\% sparsity, SCAR-LP with LP-defined supernodes achieves perplexity 10.71 and SCAR-LP with activation-L2-defined supernodes achieves perplexity 10.71 (shared protected channels: 15.4\% of each set).

\begin{figure}[H]
    \centering
    \includegraphics[width=\columnwidth]{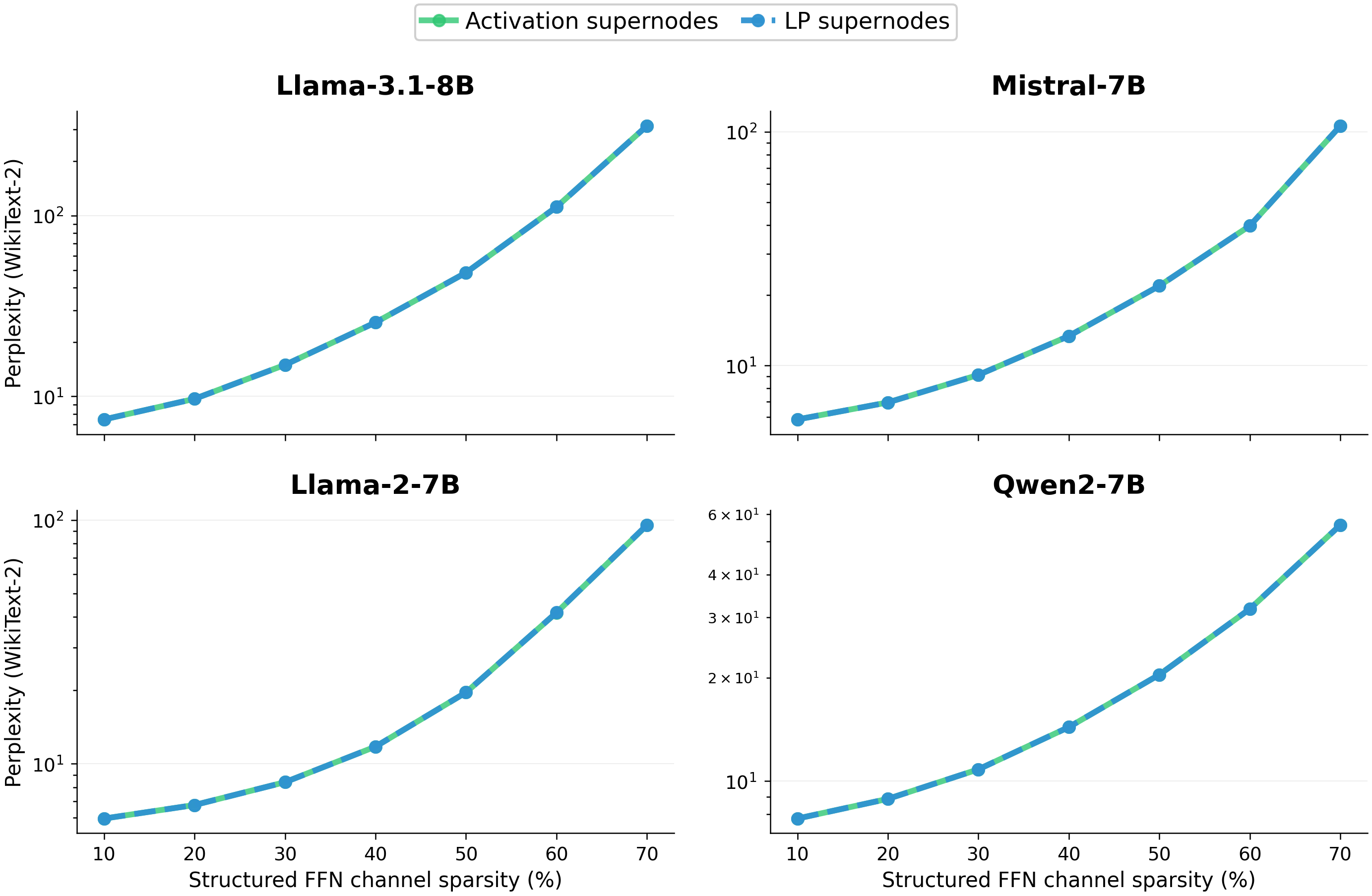}
    \caption{\textbf{Protection-set ablation across models.} Sparsity--perplexity curves for SCAR-LP with LP-defined supernodes (blue) versus activation-defined supernodes (green) on Llama-3.1-8B, Mistral-7B, Llama-2-7B, and Qwen2-7B. The two protection definitions give similar pruning curves, confirming that activation supernodes are an effective surrogate protected set even though they share only $\approx$5--10\% of channel identities with LP supernodes at 1\%.}
    \label{fig:supernode-def-sweep}
\end{figure}

Thus, activation-defined supernodes can serve as an effective \emph{surrogate protected set} within SCAR.
However, using activation magnitude as the \emph{pruning score} (rather than LP) yields much worse results: plain activation-L2 pruning gives perplexity 26.84 at the same sparsity on 70B.
Our interpretation is that LP is the more direct loss-sensitive lens for \emph{scoring} channels, while activation outliers identify structurally critical channels that are effective for \emph{protection}.
Figure~\ref{fig:supernode-def-sweep} shows the resulting pruning curves side by side across all four 7--8B models: LP-protected and activation-protected SCAR-LP curves are similar throughout the 10--70\% sparsity range, consistent with the interchangeability of the two definitions at the \emph{protection} step.

\paragraph{Factor decomposition of LP concentration.}
Figure~\ref{fig:factor-concentration} decomposes LP into its constituent factors at both 8B and 70B scale.
This figure uses a matched aggregation protocol across the two model sizes, so its 8B percentage is not the same statistic as the per-layer median reported in the abstract and main characterization.
Activation power alone captures 27.0\% (8B) and 21.1\% (70B) of top-1\% mass, and its share actually \emph{decreases} with scale.
Curvature $(s_i^2)$ alone captures only 3.2\% (8B) and 5.6\% (70B).
Factored LP (activation $\times$ curvature) captures 38.3\% (8B) and 32.1\% (70B).
But exact LP $= \frac{1}{2}\mathbb{E}[(u_i s_i)^2]$ captures 40.2\% (8B) and 61.4\% (70B), a 1.53$\times$ increase with scale.
The gap between exact and factored LP reflects within-token dependence between activation magnitude and projected-gradient magnitude; it widens from 1.8pp at 8B to 29.3pp at 70B, indicating that this dependence is a major driver of supernode concentration at scale.

\begin{figure}[H]
    \centering
    \includegraphics[width=0.82\columnwidth]{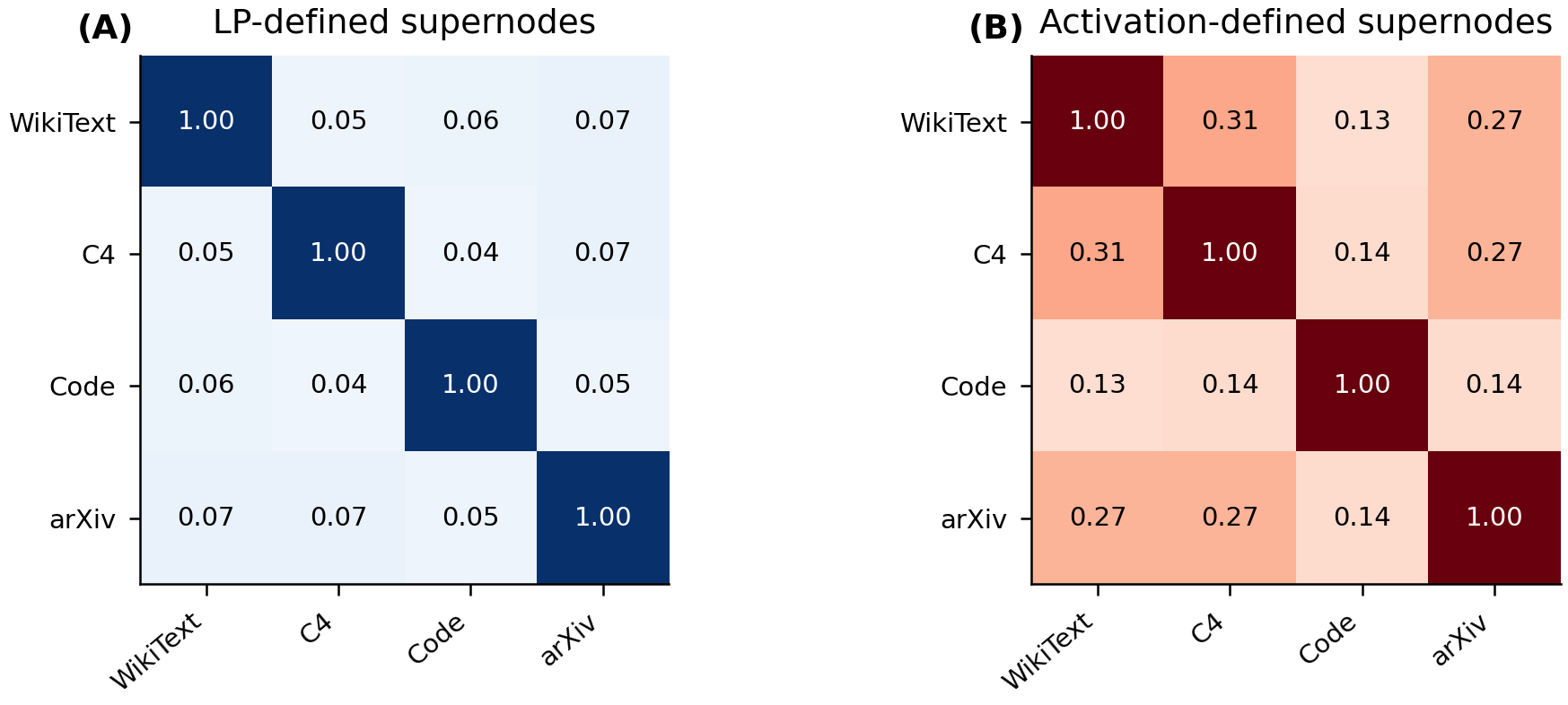}
    \caption{\textbf{Domain stability: LP vs activation supernodes.} Activation-defined top-1\% channel sets have substantially higher cross-domain Jaccard overlap than LP-defined sets; the mean off-diagonal overlap is about 3.7$\times$ larger.}
    \label{fig:supernode-stability-lp-vs-act}
\end{figure}

\paragraph{Domain stability.}
Activation supernodes are 3.7$\times$ more stable across calibration domains than LP supernodes (Fig.~\ref{fig:supernode-stability-lp-vs-act}).
This is plausible: activation outliers appear more tied to model-level or architecture-level structure and are less domain-specific than LP-defined supernodes.

\begin{figure}[H]
    \centering
    \includegraphics[width=\columnwidth]{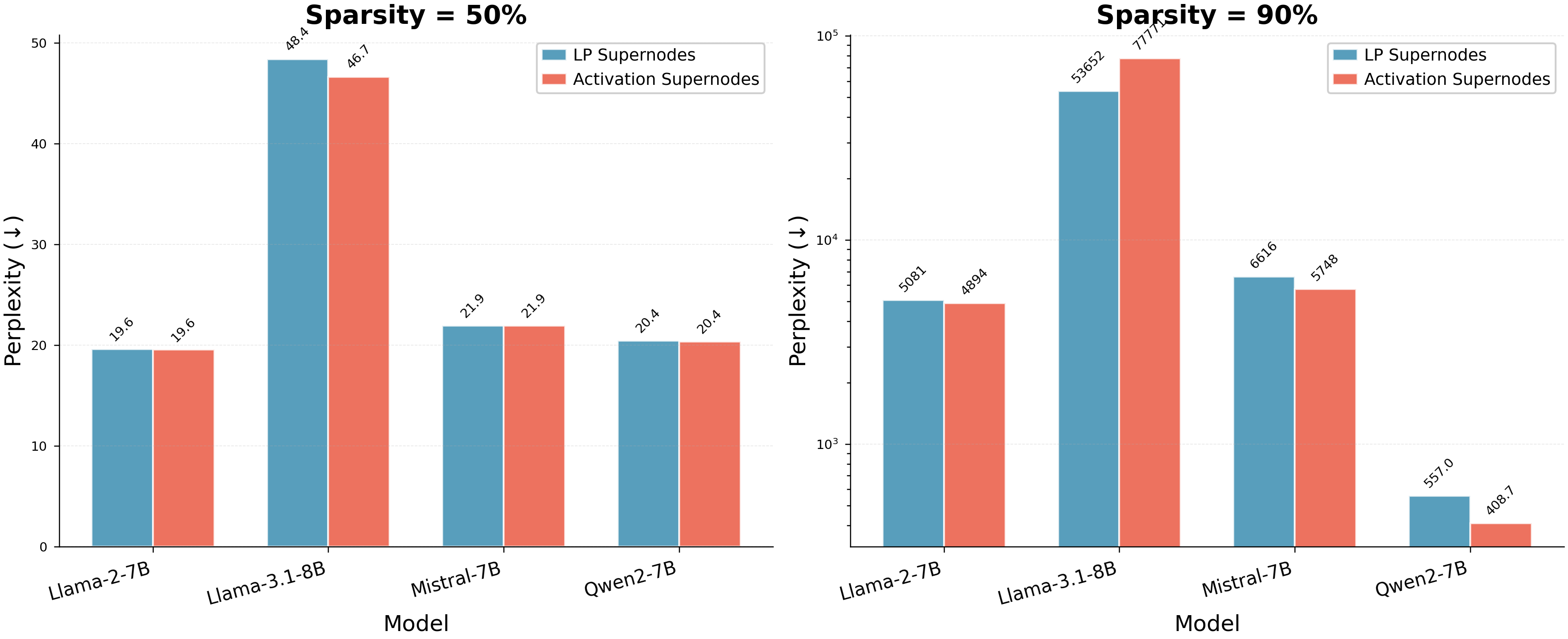}
    \caption{\textbf{LP vs activation supernode pruning.} Pruning performance is similar across models for LP-defined and activation-defined protected sets. The 50\% Llama-3.1-8B values use the 128-sequence calibration protocol from Table~\ref{tab:calibration}, so they are not directly comparable to the 64-sequence main Table~\ref{tab:main} setting. The highest-sparsity point is a stress test outside the main operating regime.}
    \label{fig:lp-vs-act-supernode-pruning}
\end{figure}

\paragraph{Pruning performance comparison.}
Despite using different supernode definitions, pruning performance is similar (Fig.~\ref{fig:lp-vs-act-supernode-pruning}).

\clearpage
\section{Cross-Model and Scale Results}
\label{app:scale-generalization}

This appendix supports the claims that LP concentration is not specific to Llama-3.1-8B and that the pruning pattern persists beyond the primary 8B setting. We group the 70B checks, the 7--8B cross-model table, and the cross-model LP-concentration summary here so the scale evidence is easy to audit in one place.

\begin{table}[H]
\centering
\caption{\textbf{Llama-3.1-70B: perplexity across sparsity levels.} Structured FFN channel pruning on WikiText-2. The gap between SCAR-LP and the channel-adapted baselines is largest at high sparsity.}
\label{tab:scale-70b}
\small
\begin{tabular}{@{}lccc@{}}
\toprule
Method & 30\% & 50\% & 70\% \\
\midrule
Unpruned & \multicolumn{3}{c}{2.79} \\
SCAR-LP & \textbf{6.23} & \textbf{10.71} & \textbf{36.29} \\
SCAR-Conn & \textbf{6.23} & 10.71 & 36.37 \\
SCAR-Prot & 6.23 & 10.77 & 38.36 \\
Wanda (channel) & 8.96 & 15.43 & 349.04 \\
Act. L2 & 8.81 & 26.84 & 420.51 \\
Magnitude (channel) & 30.22 & 123.48 & 14061.21 \\
SparseGPT (channel) & 73.00 & 224.41 & 24717.15 \\
\bottomrule
\end{tabular}

\end{table}

\begin{figure}[H]
    \centering
    \includegraphics[width=0.92\columnwidth]{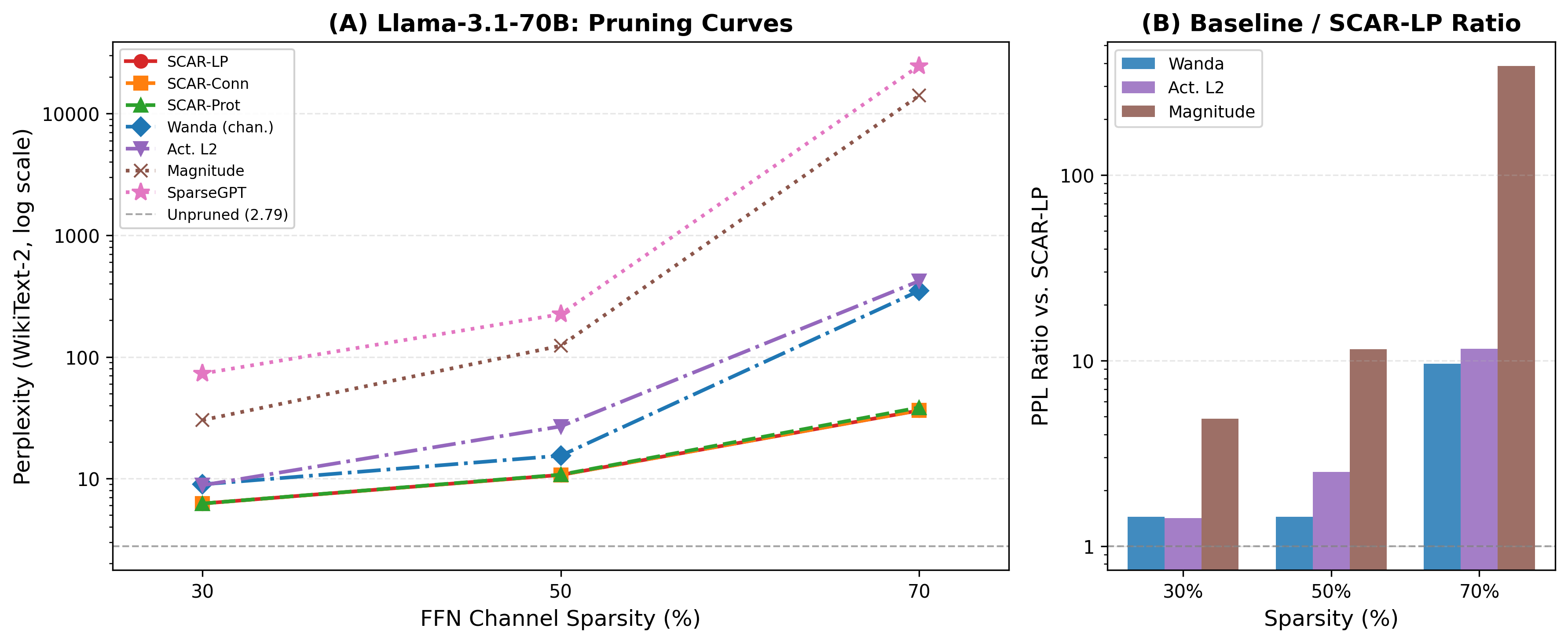}
    \caption{\textbf{70B scale validation.} \textbf{(A)} Sparsity--perplexity curves. \textbf{(B)} Baseline perplexity relative to SCAR-LP at the same sparsity.}
    \label{fig:scale-70b-curves}
\end{figure}

\begin{table}[H]
\centering
\caption{\textbf{Cross-model generalization} at 50\% sparsity. SCAR benefits remain consistent across Mistral-7B, Llama-2-7B, and Qwen2-7B.}
\label{tab:generalization}
\small
\ifdefined\NeurIPSSubmission
\begin{tabular}{@{}llccc@{}}
\toprule
Model & Method & PPL$\downarrow$ & MMLU & Avg.$\uparrow$ \\
\midrule
Mistral-7B & Wanda (channel) & 352.0 & \textbf{24.0} & 42.9 \\
Mistral-7B & SCAR-Conn & \textbf{25.3} & 22.0 & \textbf{44.1} \\
Llama-2-7B & Wanda (channel) & 62.0 & 23.0 & 44.3 \\
Llama-2-7B & SCAR-Conn & \textbf{19.5} & \textbf{24.0} & \textbf{47.4} \\
Qwen2-7B & Wanda (channel) & 89.1 & \textbf{26.0} & 39.4 \\
Qwen2-7B & SCAR-Conn & \textbf{21.2} & \textbf{26.0} & \textbf{42.0} \\
\bottomrule
\end{tabular}

\else
\resizebox{\columnwidth}{!}{
\begin{tabular}{@{}llccc@{}}
\toprule
Model & Method & PPL$\downarrow$ & MMLU & Avg.$\uparrow$ \\
\midrule
Mistral-7B & Wanda (channel) & 352.0 & \textbf{24.0} & 42.9 \\
Mistral-7B & SCAR-Conn & \textbf{25.3} & 22.0 & \textbf{44.1} \\
Llama-2-7B & Wanda (channel) & 62.0 & 23.0 & 44.3 \\
Llama-2-7B & SCAR-Conn & \textbf{19.5} & \textbf{24.0} & \textbf{47.4} \\
Qwen2-7B & Wanda (channel) & 89.1 & \textbf{26.0} & 39.4 \\
Qwen2-7B & SCAR-Conn & \textbf{21.2} & \textbf{26.0} & \textbf{42.0} \\
\bottomrule
\end{tabular}
}
\fi
\end{table}

\begin{table}[H]
\centering
\caption{\textbf{Cross-model LP concentration.} Per-layer top-1\% LP mass and LP max/mean summaries under the selected paper calibration runs.}
\label{tab:lp-concentration-models}
\small
\ifdefined\NeurIPSSubmission
\begin{tabular}{@{}lccc@{}}
\toprule
Model & Top-1\% LP mass median & Range & LP max/mean median \\
\midrule
Llama-3.1-8B & 58.7 & 33.0--86.1 & 2657 \\
Mistral-7B & 59.1 & 45.1--98.9 & 2242 \\
Llama-2-7B & 50.8 & 31.3--97.8 & 1242 \\
Qwen2-7B & 68.1 & 42.0--99.7 & 2814 \\
OLMo-2-7B & 71.7 & 33.3--94.1 & 3690 \\
\bottomrule
\end{tabular}

\else
\resizebox{\columnwidth}{!}{
\begin{tabular}{@{}lccc@{}}
\toprule
Model & Top-1\% LP mass median & Range & LP max/mean median \\
\midrule
Llama-3.1-8B & 58.7 & 33.0--86.1 & 2657 \\
Mistral-7B & 59.1 & 45.1--98.9 & 2242 \\
Llama-2-7B & 50.8 & 31.3--97.8 & 1242 \\
Qwen2-7B & 68.1 & 42.0--99.7 & 2814 \\
OLMo-2-7B & 71.7 & 33.3--94.1 & 3690 \\
\bottomrule
\end{tabular}
}
\fi
\end{table}

\begin{table}[H]
\centering
\caption{\textbf{Llama-3.1-70B at 50\% FFN sparsity: downstream tasks} (500 samples per task). SCAR variants retain the best average accuracy across eight tasks, with SCAR-LP strongest on MMLU. We still treat these task-level results as secondary to the more stable perplexity-based comparison in the main text.}
\label{tab:scale-70b-benchmarks}
{\setlength{\tabcolsep}{2pt}\scriptsize
\resizebox{\columnwidth}{!}{\begin{tabular}{@{}lcccccccccc@{}}
\toprule
Method & PPL & MMLU & HellaSwag & PIQA & BoolQ & WinoGrande & ARC-E & ARC-C & OBQA & Avg. \\
\midrule
Unpruned & 2.79 & 74.4 & 71.4 & 85.4 & 87.0 & 67.4 & 91.6 & 67.2 & 89.0 & 79.2 \\
\midrule
\textbf{SCAR-LP} & \textbf{10.71} & 43.4 & \textbf{49.4} & 73.0 & \textbf{73.4} & 53.0 & 64.8 & 37.4 & 43.0 & 54.6 \\
SCAR-Conn & \textbf{10.71} & \textbf{44.0} & 49.0 & 73.0 & 72.8 & 53.2 & 65.2 & 37.6 & \textbf{43.2} & \textbf{54.8} \\
SCAR-Prot & 10.76 & 42.4 & 48.8 & 74.0 & 72.2 & 53.2 & \textbf{65.4} & \textbf{38.0} & 43.0 & 54.6 \\
Wanda (chan.) & 15.43 & 34.2 & 49.2 & \textbf{77.2} & 69.0 & \textbf{58.0} & 60.4 & 35.4 & 40.2 & 53.0 \\
Act. L2 & 26.84 & 32.2 & 42.6 & 67.6 & 52.2 & 49.6 & 44.0 & 25.2 & 30.8 & 43.0 \\
Magnitude & 123.48 & 28.0 & 39.6 & 72.6 & 62.2 & 52.4 & 40.8 & 29.4 & 31.8 & 44.6 \\
\bottomrule
\end{tabular}
}
}
\end{table}

\clearpage
\section{OLMo-2 Pretraining Dynamics}
\label{app:olmo-training-dynamics}

\paragraph{Setup.}
Figure~\ref{fig:olmo-trajectory-main} in the main text shows the OLMo-2 trajectory. We compute \(\LP\) on ten OLMo-2-7B checkpoints spanning $\sim$1B to $\sim$4T training tokens (nine stage-1 steps plus the final post-stage-2 release, \texttt{main}). All checkpoints are evaluated with the identical calibration protocol used for Llama-3.1-8B: 64 sequences of 512 tokens from the WikiText-2 training split.

\paragraph{Final-checkpoint concentration.}
On OLMo-2-7B (\texttt{main}), the top 1\% of FFN channels by LP captures a median of \OLMoTopRhoMassFinalPct\% of layer LP mass (range over 32 layers: 33.3--94.1\%). The median LP max/mean ratio is \OLMoMaxMeanFinal{}, and the mean activation-outlier ratio is $\approx$25$\times$. This matches the FFN-channel concentration pattern we document for Llama, Mistral, and Qwen; on OLMo the top-1\% mass fraction is in fact \emph{higher} than on Llama-3.1-8B.

\paragraph{Interpretation.}
Figure~\ref{fig:olmo-trajectory-main} suggests an emergence-and-sorting account: the FFN progressively develops a small set of channels whose joint activation--gradient product dominates the loss proxy, but the specific channels filling that role are not fixed until fairly late in pretraining. We frame this as evidence rather than a formal theory; confirming the underlying mechanism (for instance whether supernodes are locked to specific massive-activation token positions, attention sinks, or residual-stream axes) is beyond this paper.

\begin{figure}[H]
    \centering
    \includegraphics[width=0.82\columnwidth]{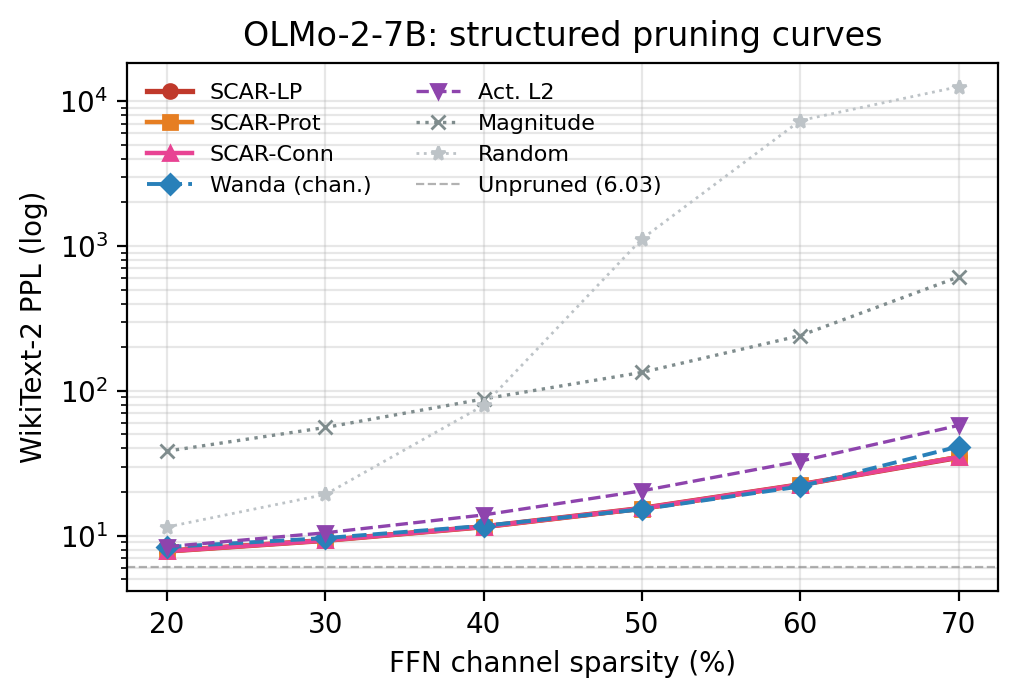}
    \caption{\textbf{OLMo-2-7B structured pruning curves.} Perplexity on WikiText-2 as a function of FFN channel sparsity for SCAR variants and baselines. SCAR-LP and Wanda-channel track each other closely across sparsities, with a small separation only at 70\%; magnitude and random pruning are far worse throughout.}
    \label{fig:olmo-pruning-curves}
\end{figure}

\paragraph{Pruning replication on OLMo-2-7B.}
To check whether core protection matters in the pruning setting for OLMo as well, we run the full structured pruning suite (SCAR-LP, SCAR-Prot, SCAR-Conn, Wanda-channel, Magnitude, activation-L2, Random) on OLMo-2-7B at six sparsity levels (20--70\%). Figure~\ref{fig:olmo-pruning-curves} shows the sparsity--perplexity curves. The unpruned perplexity is \OLMoPPLUnpruned{}; at 50\% FFN channel sparsity SCAR-LP reaches \OLMoPPLSCARLP{} versus \OLMoPPLWandaChan{} for channel-adapted Wanda, essentially tied at this sparsity with Wanda marginally lower, while magnitude pruning degrades sharply (\OLMoPPLMagnitude{}) and random pruning yields extremely high perplexity ($\gg$10\textsuperscript{3}). The qualitative pattern on OLMo differs from Llama-3.1-8B: SCAR-LP and Wanda remain close across the 20--70\% sparsity range, with Wanda rising slightly faster only at the highest sparsity (70\%). This is consistent with the broader observation that extreme Wanda degradation is most pronounced on Llama-3.1-8B, while other SwiGLU models show a much smaller SCAR vs.\ Wanda gap (see Fig.~\ref{fig:scale-70b-curves} for the 70B analogue).

\clearpage
\section{Read-Halo Diagnostics}
\label{app:halo-rescue}

This appendix details the cross-layer read-halo experiments summarized in \S\ref{sec:halos}. For each valid adjacent transition $\ell\!\to\!\ell{+}1$, we compute the supernode write support $\mathcal{S}_\ell$ and the read halo $\mathcal{R}_{\ell+1}$, defined as the top-$\eta_r$ next-layer channels by ReadConn from $\mathcal{S}_\ell$. Transition-based summaries exclude the final layer because it has no next FFN layer. All experiments use Llama-3.1-8B with the same WikiText-2 calibration as the main paper.

\paragraph{Within-set redundancy.}
We compare mean off-diagonal absolute Pearson correlation of next-layer activations $u$ within $\mathcal{R}_{\ell+1}$ versus a matched random subset of non-halo channels and a matched lowest-ReadConn non-halo subset. If read halos echo the supernode signal, within-halo activation correlation should exceed the random baseline.

\paragraph{Read-halo as prune-preferred.}
At fixed FFN sparsity $s$, we compare \SCAR{}-LP (rank by $\LP$, lowest first, supernodes protected) to \SCAR{}-PruneReadHalo, which uses
\begin{equation}
    \mathrm{Score}_j^{\mathrm{PRH}}(\alpha) = \LP_j \cdot \bigl(1 - \alpha\,\widehat{\mathrm{ReadConn}}_j\bigr),
    \label{eq:rhalo-score}
\end{equation}
where $\widehat{\mathrm{ReadConn}}_j$ is normalized to $[0,1]$ within each layer. $\alpha=0$ recovers \SCAR{}-LP; $\alpha\!>\!0$ biases toward pruning read-halo channels first. If read halos are redundant given the protected core, this variant should match or improve on \SCAR{}-LP perplexity at moderate sparsity.

\paragraph{Support-ablation response.}
We ablate $\mathcal{S}_\ell$ at the input to layer $\ell{+}1$ (post-attention layernorm output) and measure mean $|\Delta u_j|$ per next-layer channel. The decile-level top-vs-bottom ReadConn ratio of $|\Delta u_j|$ tests whether the supernode-emphasized residual coordinates causally drive the read halo.

\begin{figure}[H]
    \centering
    \includegraphics[width=0.95\columnwidth]{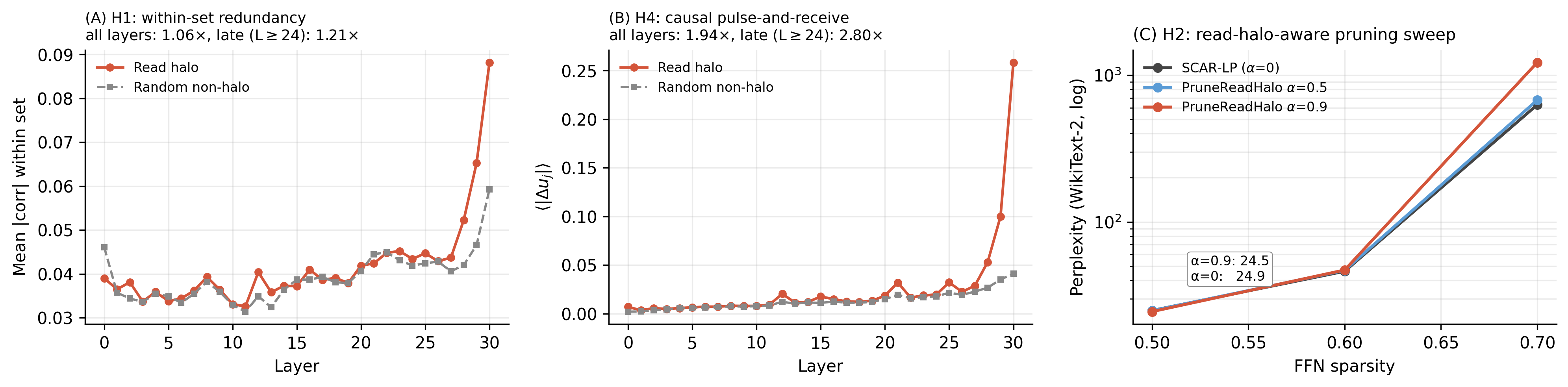}
    \caption{\textbf{Cross-layer read halos echo the supernode core.}
    \textbf{(A)} Within-set activation redundancy: mean $|{\rm corr}|$ within the read halo exceeds matched random non-halo controls; effect is modest globally ($1.06\times$) but sharpens in late layers ($1.21\times$ on L$\geq$24, peaking near 2.3$\times$ at L=30).
    \textbf{(B)} Causal pulse-and-receive: ablating supernode write support $\mathcal{S}_\ell$ at the input of layer $\ell{+}1$ produces $1.94\times$ larger $|\Delta u_j|$ on read-halo channels overall and $2.80\times$ larger on late layers; this is the cleanest causal signature of the supernode-halo coupling.
    \textbf{(C)} Read-halo-aware pruning sweep: at 50\% sparsity, biasing the prune score toward read-halo channels (\SCAR{}-PruneReadHalo, $\alpha=0.9$) gives a small improvement over LP-only pruning; the ranking flips at 70\%, where read halos become load-bearing.}
    \label{fig:halo-rescue}
\end{figure}

\paragraph{Results.}
Figure~\ref{fig:halo-rescue} summarizes the three signatures. \textbf{Panel A:} mean $|{\rm corr}|$ ratio is $1.06\times$ across valid adjacent transitions and $1.21\times$ averaged over late transitions (source L$\geq$24); the per-layer ratio peaks at $\approx$$2.3\times$ near L=30. \textbf{Panel B:} mean $\langle |\Delta u_j|\rangle$ ratio is $1.94\times$ overall and $2.80\times$ on late layers, with the late-layer single-layer maximum reaching $\approx$$6\times$. \textbf{Panel C:} at 50\% sparsity, \SCAR{}-PruneReadHalo with $\alpha=0.9$ reaches PPL $24.50$ versus $24.94$ for $\alpha=0$ (\SCAR{}-LP), a small improvement at moderate sparsity. At 60\% the two are tied; at 70\% the read-halo bias hurts (PPL $1221$ vs.\ $629$). This sparsity dependence is itself informative: read halos are partially redundant when other capacity is intact, but become load-bearing once the budget forces other channels out. Absolute PPL values in panel C come from a self-contained side experiment using the same calibration as the main pipeline.

\clearpage
\section{Read-Halo Robustness and \SCAR{}-HaloProtect}
\label{app:halo-extended}

This appendix supplies the quantitative basis for two refinements discussed in \S\ref{sec:halos}: robustness of the read-halo signal across supernode definitions, and the hard \SCAR{}-HaloProtect pruning variant. These experiments are produced by the side-experiment pipeline and are interpreted by relative HP/SCAR-LP ratios rather than by comparing absolute perplexities to the main-pipeline table.

\begin{figure}[H]
    \centering
    \includegraphics[width=\columnwidth]{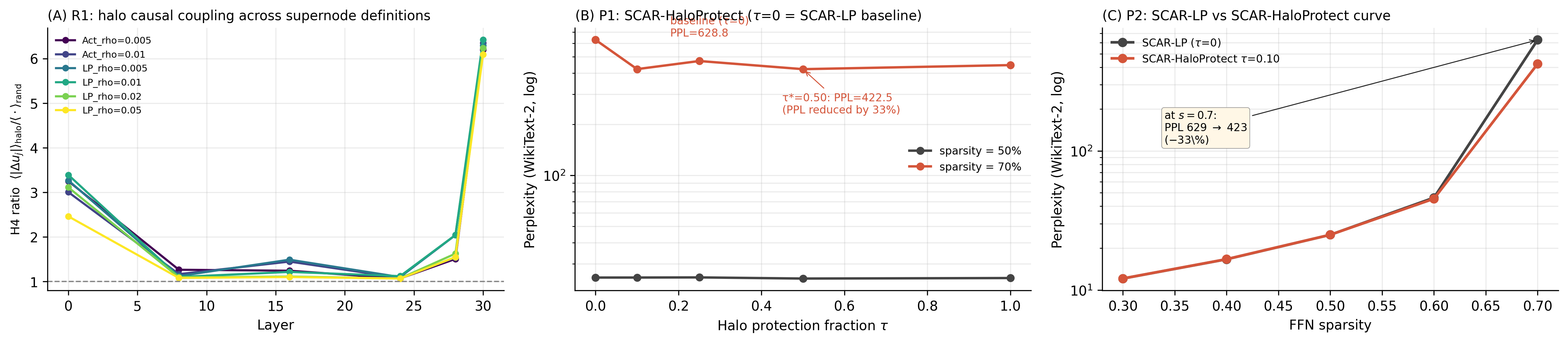}
    \caption{\textbf{Halo robustness, \SCAR{}-HaloProtect, and the sparsity-sweep curve.}
    \textbf{(A)} Halo / random ratio across supernode definitions on six probed layers. LP-defined ($\rho \in \{0.5\%, 1\%, 2\%, 5\%\}$) and activation-defined ($\rho \in \{0.5\%, 1\%\}$) supernodes all yield the same per-layer pattern; late-layer (L=30) ratio is $\approx\!6.1$--$6.4\times$ across every definition.
    \textbf{(B)} \SCAR{}-HaloProtect at fixed sparsity vs.\ halo-protection fraction $\tau$. At $s=50\%$ halo protection is essentially neutral in this side pipeline; at $s=70\%$ protecting the top $10\%$ of the read-halo set reduces perplexity from 628.8 to 422.7 (33\%), nearly matching the best point at $\tau=0.50$.
    \textbf{(C)} Sparsity-sweep curve. \SCAR{}-LP and \SCAR{}-HaloProtect track each other through $s\!=\!0.6$ but diverge sharply at $s\!=\!0.7$, exposing the moderate-vs-aggressive regime change. Halos behave like a small reserve of supernode messengers that becomes essential only under aggressive pruning.}
    \label{fig:halo-extended}
\end{figure}

\paragraph{Robustness across supernode definitions.}
For each definition we recompute $\mathcal{S}_\ell$, then $\mathcal{R}_{\ell+1}$, then re-run the support-ablation response on six representative layers ($L \in \{0, 8, 16, 24, 28, 30\}$). Figure~\ref{fig:halo-extended}A shows the resulting halo / random ratio. Across $\rho \in \{0.5\%, 1\%, 2\%, 5\%\}$ on LP and $\rho \in \{0.5\%, 1\%\}$ on activation magnitude, the per-layer pattern is essentially the same: a strong edge effect (early and late layers $>2\times$, peaking $\approx\!6.2$--$6.4\times$ at $L=30$) with modest middle-layer ratios (1.07--1.49). The mean ratio is largely insensitive to the definition (2.23--2.55) but does decrease slightly with looser thresholds, consistent with the strictest cores producing the cleanest halo signal. We treat this as evidence that the halo construction is anchored by the residual support $\mathcal{S}_\ell$ rather than by any particular threshold or scoring choice.

\paragraph{\SCAR{}-HaloProtect at fixed sparsity.}
\SCAR{}-HaloProtect protects supernodes (always) and the top $\tau$ fraction of the read-halo set in each layer; the rest is pruned by lowest LP under the same per-layer caps. We sweep $\tau \in \{0.0, 0.10, 0.25, 0.50, 1.0\}$ at sparsity $s \in \{0.50, 0.70\}$ (Fig.~\ref{fig:halo-extended}B). Numerical results:
\begin{itemize}\setlength\itemsep{1pt}
  \item $s=0.50$: $\tau$=0$\to$24.94, $\tau$=0.10$\to$24.95, $\tau$=0.25$\to$25.00, $\tau$=0.50$\to$24.61, $\tau$=1.0$\to$24.78. Halo protection is effectively neutral here.
  \item $s=0.70$: $\tau$=0$\to$\textbf{628.75}, $\tau$=0.10$\to$\textbf{422.73}, $\tau$=0.25$\to$471.89, $\tau$=0.50$\to$422.49, $\tau$=1.0$\to$446.36. Protecting just the top $10\%$ of the read-halo set cuts perplexity by $33\%$.
\end{itemize}
The number of additional channels protected at $\tau=0.10$ is about 4.5k globally, because the read-halo set is the top $\eta_r=10\%$ of channels and HaloProtect protects the top $\tau=10\%$ of that set, i.e. roughly 1\% of FFN channels over valid next-layer read-halo transitions. The flattening of the curve beyond $\tau=0.10$ at 70\% sparsity suggests that the active messenger fraction is small; the minimum is nearly tied at $\tau=0.50$, but $\tau=0.10$ already achieves the reported 33\% reduction.

\paragraph{Implementation note: hard vs.\ soft halo protection.}
The \SCAR{}-HaloProtect variant tested in this appendix uses \emph{hard} halo protection: the top $\tau$ fraction of the read-halo set is excluded from the candidate prune pool, exactly as the supernode core is. This is the operational definition implied by the messenger interpretation. A separate soft implementation option, \texttt{supernode\_read\_halo\_protect\_score}, downweights low-rank halos in a continuous score and is therefore a different algorithm; we do not use it for the experiments below. All halo-rescue P-prefixed experiments come from the self-contained pipeline in \texttt{run\_halo\_extended.py}.

\paragraph{Sparsity-sweep curve.}
We sweep the same pruning pipeline at $s \in \{0.30, 0.40, 0.50, 0.60, 0.70\}$ for $\tau \in \{0, 0.10\}$ to expose the regime change directly (Fig.~\ref{fig:halo-extended}C). \SCAR{}-LP and \SCAR{}-HaloProtect track each other tightly through $s\!=\!0.6$ -- halo protection is essentially free at moderate sparsity -- and diverge sharply at $s\!=\!0.7$, where \SCAR{}-HaloProtect retains usable perplexity while \SCAR{}-LP collapses. This is the cleanest expression of the load-bearing finding: read halos are a small reserve of channels whose role becomes decisive only when the budget pushes the network past the point where the supernode core can carry the message alone.

\paragraph{Cross-model behavior and the messenger--collapse correlation.}
We replicate the sparsity-sweep curve on Mistral-7B, Llama-2-7B, Qwen2-7B, and Llama-3.1-70B (Fig.~\ref{fig:cross-model-curves}) using the same calibration protocol and supernode JSON pipeline. The pattern is consistent across all five models: \SCAR{}-LP and \SCAR{}-HaloProtect track each other at moderate sparsity, with a positive HaloProtect benefit emerging at $s=0.7$ that is \emph{largest where SCAR-LP collapses most}. Numerical 70\% perplexities (SCAR-LP / HaloProtect / reduction): Llama-3.1-8B 628.8 / 422.7 / 33\%, Mistral-7B 117.0 / 95.0 / 19\%, Qwen2-7B 66.4 / 62.8 / 5\%, Llama-2-7B 106.4 / 105.5 / 1\%, Llama-3.1-70B 36.5 / 35.9 / 2\%.

The messenger--collapse correlation compares HaloProtect rescue at $s\!=\!70\%$ against the baseline collapse ratio $\mathrm{PPL}_{70\%}/\mathrm{PPL}_{\mathrm{low}}$, where \(\mathrm{PPL}_{\mathrm{low}}\) is the side-pipeline \SCAR{}-LP perplexity at the lowest evaluated sparsity below 50\% (\(s=30\%\)). This gives Spearman $\rho=0.70$ and log-Pearson $r=0.86$ across five models (Llama-3.1-8B, Mistral-7B, Llama-2-7B, Qwen2-7B, Llama-3.1-70B). Models whose unprotected pruning collapses more are the ones for which protecting halos tends to rescue more. Absolute perplexities in the side experiment differ from main-text Table~\ref{tab:main} because the side experiment uses a self-contained pruning loop without the full main-pipeline post-processing; the rescue ratio re-anchored onto the main-pipeline scale (Fig.~\ref{fig:cross-model-curves}) is the meaningful object. With $n\!=\!5$ models, the correlation should be read as a consistency check on the messenger picture rather than a precise scaling law.

\end{document}